\documentclass[runningheads]{llncs}

\usepackage{eccv}

\usepackage{eccvabbrv}

\usepackage{graphicx}
\usepackage{booktabs}

\usepackage[accsupp]{axessibility}  

\usepackage{hyperref}

\usepackage{orcidlink}

\usepackage{multirow}
\usepackage[export]{adjustbox}
\usepackage{algorithm}
\usepackage{algpseudocode}
\usepackage{pifont}
\usepackage{wrapfig}

\newif\ifshowcomments
\showcommentstrue

\ifshowcomments

\newcommand{\TODO}[1]{{\color{red}{[TODO: #1]}}}

\newcommand{\revised}[1]{{\color[rgb]{0.2,0.7,0.2}{#1}}}

\else
\newcommand{\TODO}[1]{}
\newcommand{\revised}[1]{}
\fi

\begin{document}

\title{Towards Real-World Adverse Weather Image Restoration: Enhancing Clearness and Semantics with Vision-Language Models} 

\titlerunning{Towards Real-World Adverse Weather Image Restoration}

\author{
Jiaqi Xu\inst{1} \and
Mengyang Wu\inst{1} \and
Xiaowei Hu\inst{2,}\thanks{Corresponding author (huxiaowei@pjlab.org.cn)} \and \\
Chi-Wing Fu\inst{1} \and
Qi Dou\inst{1} \and
Pheng-Ann Heng\inst{1}}
\authorrunning{J.~Xu et al.}
\institute{The Chinese University of Hong Kong \and
Shanghai Artificial Intelligence Laboratory}
\maketitle

\begin{abstract}

This paper addresses the limitations of adverse weather image restoration approaches trained on synthetic data when applied to real-world scenarios.
We formulate a semi-supervised learning framework employing vision-language models to enhance restoration performance across diverse adverse weather conditions in real-world settings.
Our approach involves assessing image clearness and providing semantics using vision-language models on real data, serving as supervision signals for training restoration models.
For clearness enhancement, we use real-world data, utilizing a dual-step strategy with pseudo-labels assessed by vision-language models and weather prompt learning.
For semantic enhancement, we integrate real-world data by adjusting weather conditions in vision-language model descriptions while preserving semantic meaning.
Additionally, we introduce an effective training strategy to bootstrap restoration performance.
Our approach achieves superior results in real-world adverse weather image restoration, demonstrated through qualitative and quantitative comparisons with state-of-the-art works.

\end{abstract}

\section{Introduction}
\label{sec:intro}

Images captured under challenging weather conditions, such as rain, haze, and snow, are plagued by a variety of artifacts that significantly affect the image quality. 
These imperfections severely impair the efficacy of outdoor vision systems.
%
Previous research efforts~\cite{fu2017removing,jiang2020multi,cai2016dehazenet,qin2020ffa,liu2018desnownet} have primarily focused on developing specialized techniques for mitigating the effects of individual weather phenomena, tailoring their models to the unique characteristics of rain, haze, or snow. 
More recently, all-in-one adverse weather removal works~\cite{li2020all,valanarasu2022transweather,chen2022learning,ozdenizci2023restoring} design single model-based methods to restore images captured under multiple adverse weather conditions. 
Despite the encouraging outcomes demonstrated on synthetic datasets by these approaches, their applicability to real-world scenarios remains notably constrained.

The limited generalization capability in real-world adverse weather images can be attributed to two main factors. 
Firstly, adverse weather removal methods are predominantly trained on synthetic datasets~\cite{qian2018attentive,liu2018desnownet,li2019heavy,fu2017removing,li2018benchmarking,chen2021all}, resulting in a domain gap when applied to real-world situations. Secondly, these methods primarily focus on restoring the visual clarity of images, often neglecting the semantic context of the scenes they depict.
Consequently, current weather removal approaches struggle with real-world data and offer marginal enhancements to downstream high-level vision tasks under adverse weather conditions.

\begin{figure}[tp]
    \centering
    \includegraphics[width=1.0\hsize]{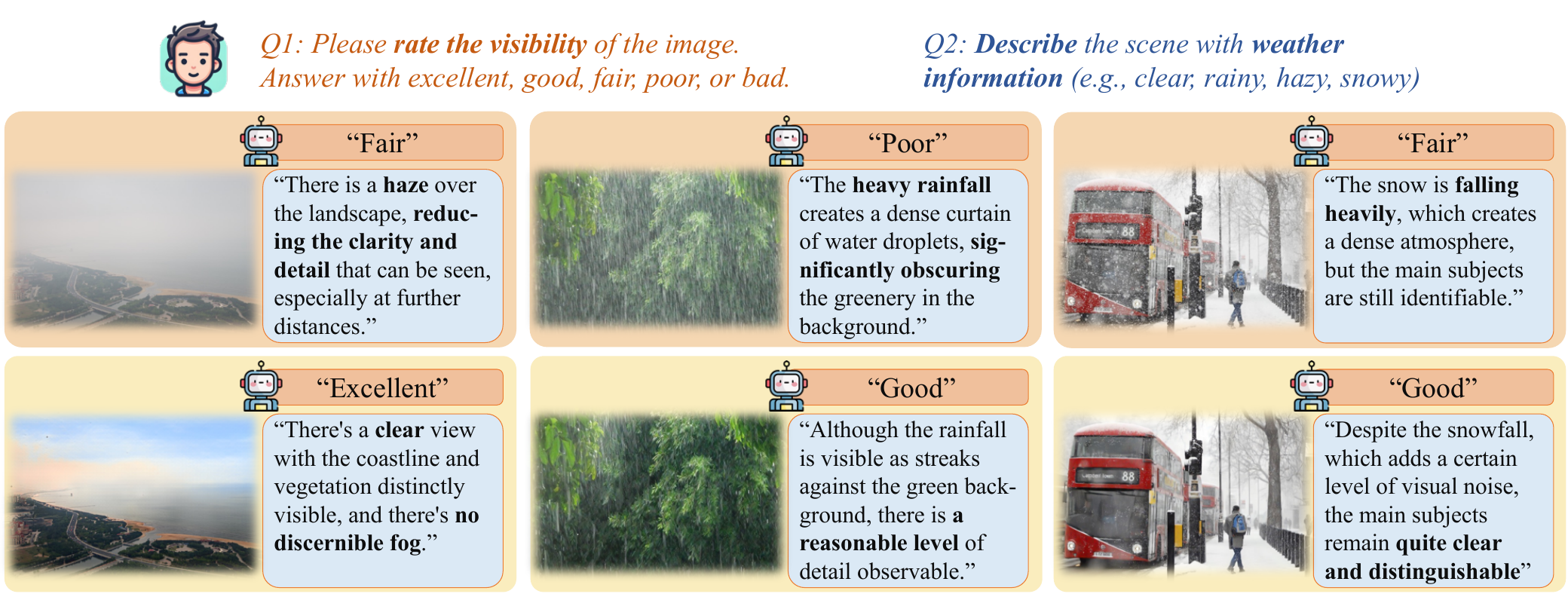}
    \caption{The \textbf{clearness} level and the \textbf{semantics} information of real-world adverse weather images are provided by large vision-language models. This assistance is instrumental in training image restoration models to effectively utilize real-world data.}
    \label{fig:intro}
\end{figure}

In response to these challenges, this work introduces a novel semi-supervised learning framework, WResVLM, that explores vision-language models (VLMs) \cite{liu2023visual,radford2021learning} to enhance image restoration in real-world scenarios across diverse adverse weather conditions.
The real-world images with the weather-related artifacts are used as the unlabeled (unpaired) data to train image restoration models and the supervision signals are provided by the large vision-language models.
As depicted in \cref{fig:intro}, large VLMs play a crucial role in assessing the \emph{clearness} levels and providing \emph{semantics} information of images under adverse weather conditions. 
This capability proves instrumental in training image restoration models effectively, enabling them to handle the complexities of real-world data.

To enhance the \textit{clearness} of the restored images produced by the restoration model, we utilize real-world data for model training, evaluating image clarity with the assistance of large vision-language models. These models, exposed to a diverse array of weather conditions during training, demonstrate proficiency in recognizing and distinguishing various weather-related scenes.
The approach involves two key steps: initially, the vision-language model is employed to assess images and select pseudo-labels for training the restoration model.
Subsequently, weather prompt learning is introduced to tailor the VLM, ultimately utilizing it to modulate the image restoration process.
This dual-step strategy enhances the restoration model's ability to address the real-world weather complexities and improve the overall clearness of the restored images.

To enhance the \textit{semantics} of the restored images, we further integrate real-world data into the model training.
This involves utilizing descriptions generated by vision-language models associated with each image, providing rich semantic information about the scene and adverse weather conditions.
A unique aspect of our method involves adjusting the weather clues in the descriptions while maintaining the semantic meaning unchanged.
This enables the training of the image restoration model to specifically target the removal of weather-related artifacts without altering the image's underlying semantics.
In contrast to methods that might overlook semantic cues, our framework incorporates vision-language models to encompass both visual clarity and semantic context, thus presenting a more comprehensive strategy for adverse weather image restoration.

Lastly, we develop a training strategy aimed at achieving effective pseudo-label initialization and iterative updates, with the primary goal of improving restoration outcomes.
We conduct experiments using real-world images captured under diverse adverse weather conditions. 
The results demonstrate that our method significantly surpasses both state-of-the-art adverse weather image restoration approaches and general image restoration methods.
Code and data are available at \href{https://github.com/jiaqixuac/WResVLM}{GitHub}.
%

%


\section{Related Work}
\label{sec:background}

\subsection{Image Restoration in Adverse Weather Conditions}
Previous works focus on restoring images captured under specific weather conditions, including deraining~\cite{fu2017removing,yang2017deep,hu2019depth,zhu2020learning,hu2021single,xiao2022image}, dehazing~\cite{he2010single,cai2016dehazenet,deng2019deep,song2023vision}, and desnowing~\cite{liu2018desnownet}.
Recent works~\cite{li2020all,valanarasu2022transweather,chen2022learning,ozdenizci2023restoring,zhu2023learning,wang2023smartassign,ye2023adverse,patil2023multi} focus on all-in-one adverse weather removal, which restores images captured under various weather conditions using a single model.
The pioneering All-in-One~\cite{li2020all} achieves this by using joint training and a unified set of model weights.
TransWeather~\cite{valanarasu2022transweather} introduces a transformer-based architecture while Chen~\etal~\cite{chen2022learning} leverage knowledge distillation and contrastive learning.
WeatherDiff~\cite{ozdenizci2023restoring} adapts the diffusion model for adverse weather artifact removal.
Zhu~\etal~\cite{zhu2023learning} learn weather-general and weather-specific features through multiple sets of model weights.
AWRCP~\cite{ye2023adverse} enhances image restoration by exploring high-quality codebook priors.
Domain adaptation technique is also utilized to handle mixed weather conditions~\cite{patil2023multi}.
More recent works explore prompting~\cite{potlapalli2023promptir}, textual information~\cite{luo2024controlling}, and customizing pre-trained diffusion models~\cite{jiang2023autodir}.
PromptIR~\cite{potlapalli2023promptir} enhances the all-in-one restoration by predicting degradation-conditioned prompts.
DA-CLIP~\cite{luo2024controlling} learns the degradation information through image-text contrastive learning.

The prior approaches typically rely on paired synthetic data~\cite{fu2017removing,qian2018attentive,liu2018desnownet,li2018benchmarking,li2019heavy,chen2021all,xu2023video} for training and evaluation, demonstrating promising results in synthetic benchmarks.
However, the trained models exhibit limited generalization capabilities toward complex real-world scenarios due to the domain gap.
Additionally, WeatherStream~\cite{zhang2023weatherstream} attempts to compile a dataset of real degenerated images with corresponding ground truth,
yet suffering from low image quality issues, \eg, compression artifacts. 


\subsection{Vision-Language Models}
Vision-language models merge computer vision and natural language processing.
%
CLIP~\cite{radford2021learning} pioneered text and image alignment through large-scale pre-training.
CLIP's versatility is also demonstrated in image manipulation fields, including backlit image enhancement~\cite{liang2023iterative} and novel concept generation~\cite{richardson2023conceptlab}.
Recent advances, including GPT-4~\cite{achiam2023gpt} and Llama~\cite{touvron2023llama}, demonstrate impressive conversational abilities.
Large VLMs like LLaVA~\cite{liu2023visual} excel in high-level multimodal visual question answering.
Recent works reveal that vision-language models are also applicable to low-level applications.
CLIP-IQA~\cite{wang2023exploring} and LIQE~\cite{zhang2023blind} expand upon the CLIP-like architecture for image quality assessment, showcasing VLMs' adaptability to technical evaluations of image quality. 
Q-Bench~\cite{wu2024qbench} highlights VLMs' inherent low-level perceptual capabilities.
These works, however, focus primarily on general technical image quality assessment and show limited abilities to help image restoration under adverse weather conditions.

\section{Methodology}
\label{sec:method}

In this work, we introduce a novel semi-supervised learning framework for all-in-one adverse weather image restoration, leveraging both labeled synthetic images and unlabeled real images.
Our motivation emphasizes the necessity to improve image restoration in the real world.
Current approaches, mostly trained on synthetic images, struggle with generalization when handling real-world adverse weather images.
They frequently overlook the image context related to weather-related artifacts in real data, resulting in their limited effectiveness.

\begin{figure}[tp]
    \centering
    \includegraphics[width=0.95\hsize]{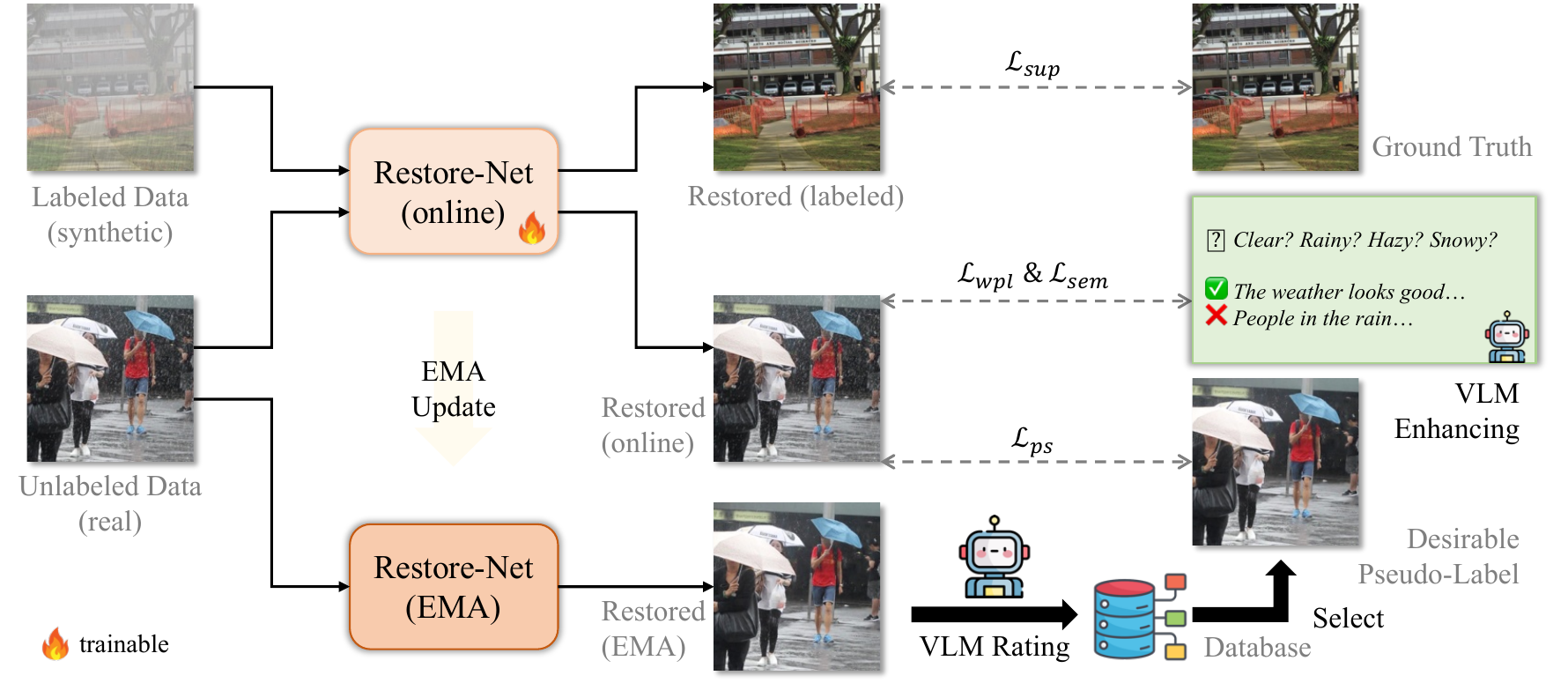}
	\caption{The schematic illustration of the proposed semi-supervised learning framework enhancing real-world image restoration by improving clearness and semantics in varied adverse weather conditions through the utilization of vision-language models.}
    \label{fig:overview}
\end{figure}


\Cref{fig:overview} shows the overall pipeline of our proposed semi-supervised learning framework for all-in-one adverse weather image restoration in real-world situations. This framework adopts several VLMs to improve the images' Clearness and Semantics during the removal of weather-related artifacts.


\subsection{Enhancing Image Clearness through Vision-Language Models}
\label{sec:clearness}

Restoring images in adverse weather conditions involves eliminating weather-related artifacts to generate ``clean'' images.
Attaining \textit{clearness} is a primary goal in adverse weather image restoration, especially in the real world.
%
In the absence of ground truth (clean) images for real-world data, the main challenge lies in determining the quality of restored images.
Moreover, limited learning objectives are designed to enhance image clearness under weather-related conditions.

Large vision-language models, trained on diverse data and vast weather imagery, exhibit strong representation abilities for image quality assessment.
%
Additionally, with the help of prompt learning, the VLMs can better distinguish well-restored images from those degraded by rain, haze, or snow.
To achieve this, we suggest two steps.
First, we employ the large vision-language models to assess the images and provide pseudo-labels for training the restoration model.
Then, we introduce weather prompt learning to empower the VLM's ability to identify clearness, ultimately utilizing it for modulating image restoration.

\subsubsection{Image Assessment and Pseudo-Labeling.}
\label{subsubsec:vlmiqa}

Our goal is to improve the restoration of real adverse weather images using unlabeled data. 
This involves training restoration models with pseudo-labels generated from the unlabeled images. 
To ensure high-quality pseudo-labels for the subsequent model training, we establish a pseudo-label database, utilizing the zero-shot capability of large vision-language models to assess adverse weather image restoration.
%

\paragraph{Image assessment.}
Given the real adverse weather images and the corresponding predictions from deweathering methods, a critical issue is to measure the image quality of the restored images.
Existing methods~\cite{talebi2018nima,ke2021musiq} for low-level image quality assessment focus mainly on technical distortions, including noise, blur, and compression artifacts.
There, however, exists a situation where an image that suffers from adverse weather is of ``good'' image quality, with little common noise, yet the visibility is largely degraded due to rain, haze, and snow.
Hence, it is imperative to find an effective way to automatically evaluate the image quality in the context of adverse weather artifact removal.

\begin{figure}[tp]
    \centering
    \includegraphics[width=\hsize]{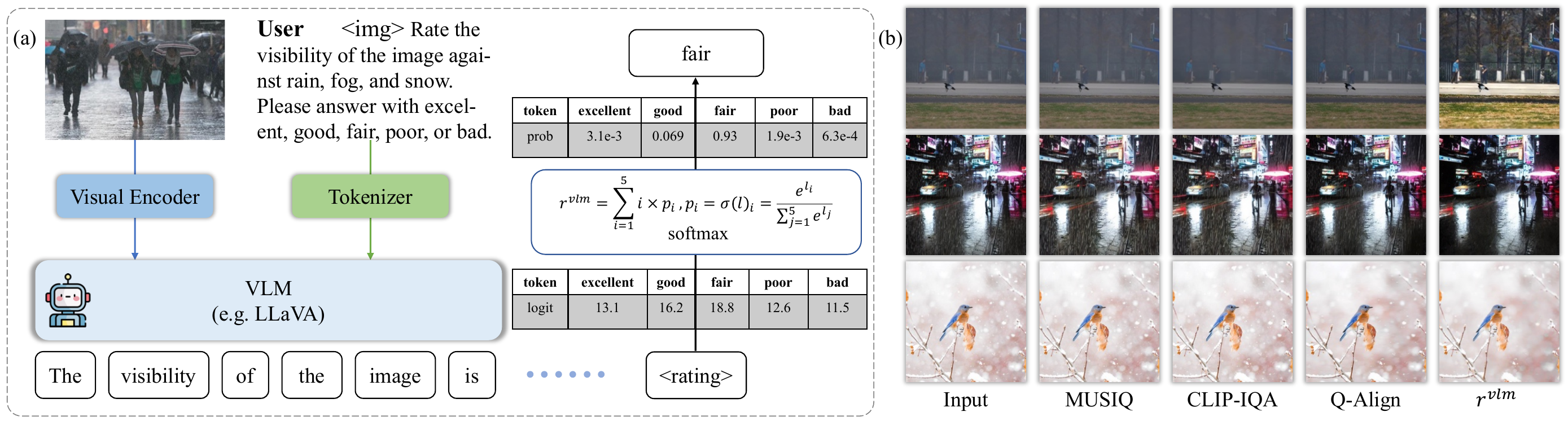}
    \caption{
    VLM-based assessment of images restored from weather-related artifacts.
    In (a), we show the process to compute the VLM's image assessment ratings $r^{vlm}$.
    In (b), we find that $r^{vlm}$ can select pseudo-labels with fewer weather-related artifacts.
    }
    \label{fig:vlmrating}
\end{figure}

Inspired by recent works~\cite{wu2024qbench,wu2024qalign} that vision-language models perform the zero-shot image quality assessment with appropriate prompting, we present to uncover the potential of VLMs for assessing the adverse weather image restoration.
Technically, we prompt the VLMs with weather-related image quality questions and convert the VLMs' responses into numerical scores.
In detail, we first design the conversion templates for enquiring about the VLM responses to assess the image as illustrated in \cref{fig:vlmrating}~(a).

Then, we adopt the commonly used five-scale ratings in the mean opinion score (MOS) studies, \ie, excellent, good, fair, poor, and bad, which correspond to the scores between one and five.
After that, we calculate the VLM-based rating $r^{vlm}$ by converting the VLMs' predicted probabilities over these five-word tokens into numerical scores using \textit{softmax}:
\begin{equation}
    r^{vlm} = \sum\nolimits_{i=1}^5 i \times p_{i}, \quad p_{i} = \sigma(l)_i = \frac{e^{l_i}}{\sum_{j=1}^5 e^{l_j}} \ ,
\end{equation}
%
where $p_i$ denotes the probability for rating $i \in \{1,2,3,4,5\}$, $l_i$ denotes the logit extracted from the language model for rating token $i$, and $\sigma$ is the $\textit{softmax}$ operation.
Thus, we obtain the visibility assessment for each restored image.

\paragraph{Pseudo-labeling.}
For the unlabeled real adverse weather image set $\mathcal{D}^u$, we assign and update the pseudo-labels $\mathcal{D}^{ps}=\{(x_i^u,y_i^{ps}) | x_i^u \in \mathcal{D}^u\}_{i=1}^M$ with desirable artifact-free pseudo-label images $y_i^{ps}$ based on the VLM-based image assessment.
Through investigation, we observe that $r^{vlm}$ is able to acquire better pseudo-labels with fewer weather-related artifacts, as shown in \cref{fig:vlmrating}~(b).

Initially, a pseudo-label database is constructed to store the current optimal pseudo-labels for the unlabeled images. 
Subsequently, throughout the model training process, we evaluate the VLM-based image visibility rating score for both the model's prediction and the recorded pseudo-labels. If the model achieves a superior restoration, we update the pseudo-label database accordingly~\cite{huang2023contrastive}. 
In practice, we use predictions from the teacher model~\cite{tarvainen2017mean} for comparison, which is an exponential moving average of the student model.
Lastly, we use the updated pseudo-labels to compute the pseudo-label loss for the online model:
\begin{equation}
    \mathcal{L}_{ps} = \mathcal{L}_{app}(\hat{y}_i, y_i^{ps}) \ ,
\end{equation}
where $\hat{y}_i$ and $y_i^{ps}$ are the prediction and the corresponding pseudo-label, respectively, and $\mathcal{L}_{app}$ is any kind of appearance loss, \eg, $\mathcal{L}_{1}$ as adopted.

\subsubsection{Weather Prompt Learning.}
\label{subsec:promptlearning}

\begin{figure}[tp]
    \centering
    \includegraphics[width=1.0\hsize]{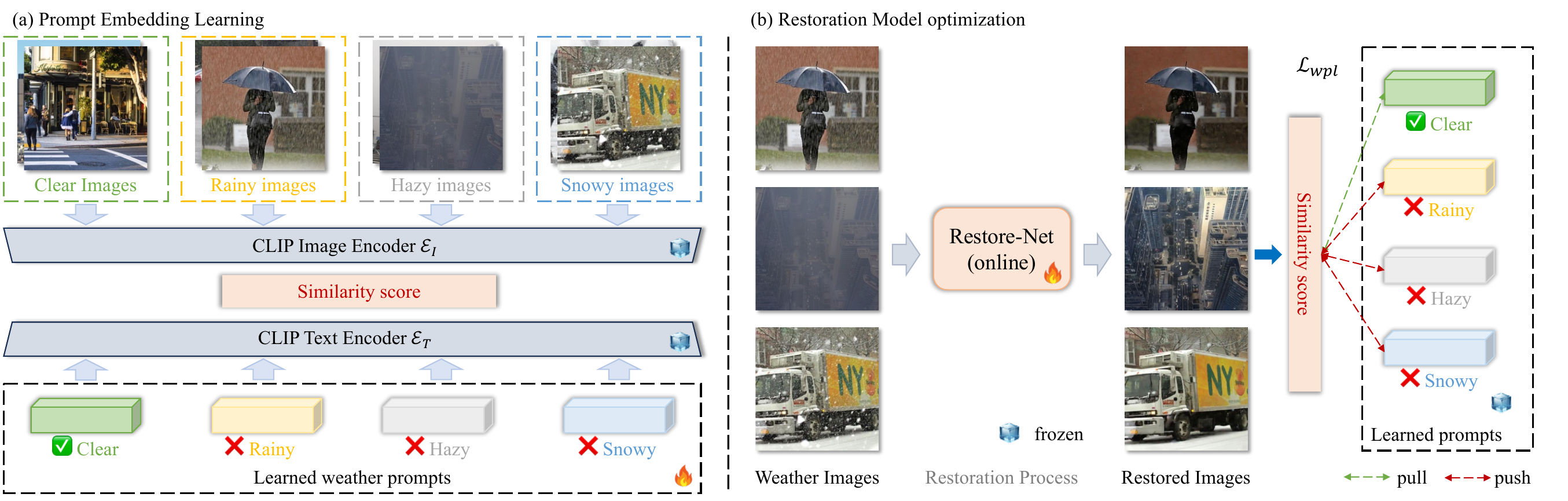}
	\caption{The workflow of the weather prompt learning approach.}
    \label{fig:promptlearning}
\end{figure}

We delve into the extensive knowledge embedded in the pre-trained vision-language model, capable of understanding the concept of images in both normal and adverse weather conditions.
Specifically, we anticipate the CLIP~\cite{radford2021learning} model to be indicator aware of image weather conditions, such as clear, rainy, hazy, or snowy. 
Subsequently, we leverage the learned concept of ``clearness'' to modulate the model's learning toward achieving clear restoration.
To enhance CLIP's ability to accurately differentiate weather in diverse scenarios, we employ the prompt learning approach to acquire prompt embeddings tailored to the image characteristics of each weather situation.
The Weather Prompt Learning process consists of two stages: the prompt embedding learning stage and the restoration model optimization stage; see \cref{fig:promptlearning}. 

\paragraph{Prompt embedding learning.}
CLIP aligns images and text within a shared feature space.
Rather than relying on fragile prompt engineering involving hand-crafted text prompts such as ``rainy'' or ``a rainy photo'', we adopt the prompt learning approach~\cite{zhou2022learning,liang2023iterative}.
Specifically, keeping the pre-trained CLIP model parameters fixed, we employ a set of four weather prompts ${t_c, t_r, t_h, t_s}$ representing clear, rain, haze, and snow conditions as learnable vectors.
%

The weather prompts $\mathcal{E}_T(t)$ are initialized in the embedding space of the CLIP's text encoder.
Meanwhile, the real images $x$ in such clear, rainy, hazy, and snowy situations are collected, which are used to extract the reference image embeddings $\mathcal{E}_I(x)$ through the CLIP's image encoder.
During the prompt embedding learning stage, the training objective is to minimize the classification loss, \ie, the cross-entropy loss, by categorizing the weather prompts into their respective weather categories $c$:
$p(c=i|x) = \sigma ( z_i ), z_i = cos(\mathcal{E}_I(x), \mathcal{E}_T(t_i))$, where $\sigma$ denotes \textit{softmax}, and $cos(\cdot,\cdot)$ denotes cosine similarity.
%
Note that the learnable weather prompt embeddings are the only parameters to be optimized during this stage; see \cref{fig:promptlearning} (a).

\paragraph{Restoration model optimization.}
With the acquired knowledge from the learned weather prompts, we direct the training of the restoration model to generate images with enhanced clearness.
Formally, during the restoration model optimization, the weather prompt learning loss $\mathcal{L}_{wpl}$ maximizes the similarity between the image embedding $\mathcal{E}_I(\hat{y})$ of the model's restored image $\hat{y}$ and the text embedding of the clear weather prompt $\mathcal{E}_T(t_c)$:
\begin{equation}
    \mathcal{L}_{wpl} = \frac {e^{cos(\mathcal{E}_I(\hat{y}), \mathcal{E}_T(t_c))}} {\sum_{t \in \{t_c,t_r,t_h,t_s\}}e^{cos(\mathcal{E}_I(\hat{y}), \mathcal{E}_T(t))}} \ .
\end{equation}

In initial investigations, employing only $\mathcal{L}_{wpl}$ optimizes the model's prediction to reduce weather-related artifacts, yet the resulting image exhibits noticeable noise. 
We hypothesize that there is room for plausible solutions within the space of minimizing the weather prompt learning loss.
To address this issue and regularize the model learning, a feature similarity loss is employed to align the model's prediction with both the pseudo-label $y^{ps}$ and the input $x^u$:
\begin{equation}
    \mathcal{L}_{feat} = \frac{1}{HW} \sum\nolimits_{i=1}^{HW} \left (1 - cos(\hat{g}_i,g_i^*) \right ) \ ,
\end{equation}
where $\hat{g},g^*$ are image features of $\hat{y}$ as well as $y^{ps}, x^u$ extracted from a pre-trained model, and $H,W$ denote the spatial dimension of the feature space.
In practice, we adopt the visual encoder of Depth Anything~\cite{yang2024depth} for feature extraction because of its robustness against various scenarios.

\subsection{Enhancing Image Semantics through Vision-Language Models}
\label{sec:semantics}

Restoring images in adverse weather conditions entails not only improving image clarity but also restoring the \textit{semantics} distorted by weather-related artifacts. 
This contributes to the effectiveness of downstream vision tasks.
%
The potential to recover image semantics is frequently disregarded by existing works trained on synthetic data.
In this work, we introduce a method that leverages the image-text understanding capability of large vision-language models to enhance image semantics in the context of adverse weather image restoration.

\subsubsection{Description-assisted semantic enhancement.}

\begin{figure}[tp]
    \centering
    \includegraphics[width=\hsize]{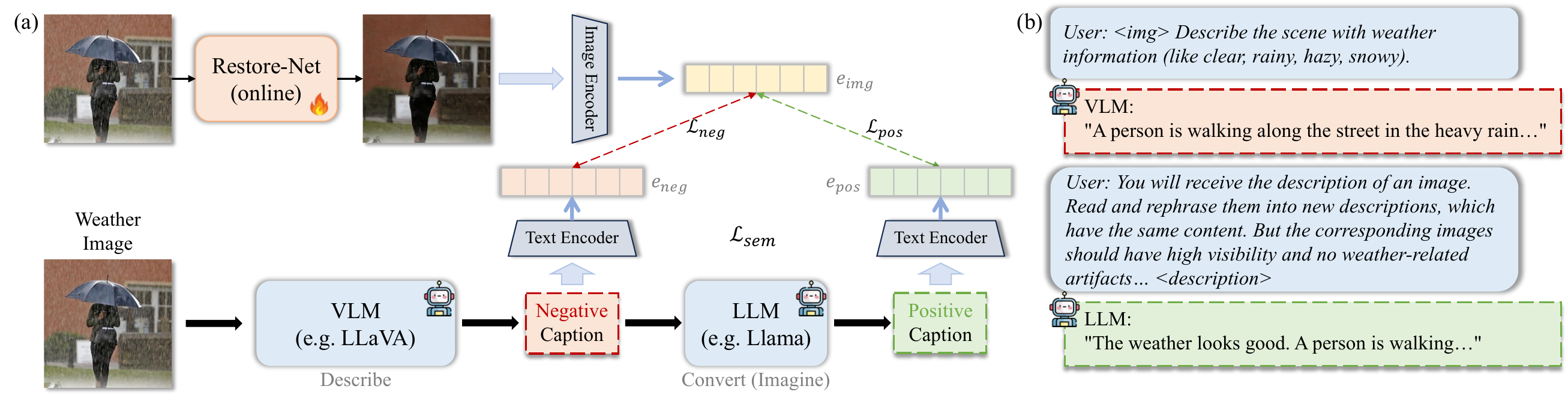}
	\caption{Description-assisted semantic enhancement with the vision-language models.}
    \label{fig:captionreg}
\end{figure}
Restoring degraded images is inherently challenging; describing their weather-affected appearance with natural language is straightforward.
Our approach uses vision-language models to generate semantic descriptions of adverse weather images, capturing both scene context and weather conditions, including degradation levels.
%
%
The comprehensive workflow is illustrated in \cref{fig:captionreg}.
Given the input image, we employ a VLM, \eg, LLaVA~\cite{liu2023visual}, to generate the (negative) caption with the weather description.
For instance, \textit{``A person is walking along the street in the heavy rain ...''} describes a scene with the object \textit{person} in the weather of \textit{rain}.
The description also provides additional environment context, like \textit{street}.

Next, we transform negative scene description $d_{neg}$ associated with the degraded image in adverse weather conditions into a pseudo-clear representation.
This transformation is achieved by prompting large language models, \eg, Llama~\cite{touvron2023llama}, to generate positive description $d_{pos}$ corresponding to the restored image.
Given the above negative description of the adverse weather, we can imagine its positive, clearly restored image, \eg, \textit{``The weather looks good. A person is walking ...''}
Intuitively, $d_{pos}$ and $d_{neg}$ should have similar descriptions of the image content, like object and environment, but dissimilar descriptions of the weather and visibility, \ie, good versus bad weather.
Unlike the weather prompt in \ref{subsec:promptlearning}, $d_{pos},d_{neg}$ are tailored to a specific image; see \cref{fig:captionreg}.

\subsubsection{Loss function.} The model training incorporates semantic-aware regularization, promoting predictions that align with positive descriptions indicative of good weather conditions.
Given the positive and negative descriptions, we formulate a description-assisted semantics regularization loss $\mathcal{L}_{sem}$:
\begin{equation}
    \mathcal{L}_{sem} = \frac {e^{cos(\mathcal{E}_I(\hat{y}), \mathcal{E}_T(d_{pos}))}} {\sum_{d \in \{d_{pos},d_{neg}\}}e^{cos(\mathcal{E}_I(\hat{y}), \mathcal{E}_T(d))}} \ .
\end{equation}
In initial trials, we observed that LLMs occasionally struggle with generating weather-varying descriptions that are content-invariant. To address this issue, we manually label certain negative-to-positive description conversions and introduce these examples in the in-context learning approach~\cite{mann2020language}.

Finally, the overall loss is a weighted combination of the supervised appearance loss $\mathcal{L}_{sup}$, semi-supervised pseudo-label loss $\mathcal{L}_{ps}$, weather prompt loss $\mathcal{L}_{wpl}$, description-assisted semantic loss $\mathcal{L}_{sem}$, and feature similarity loss $\mathcal{L}_{feat}$:
\begin{equation}
    \mathcal{L} = \mathcal{L}_{sup} + w_1 \times \mathcal{L}_{ps} + w_2 \times \mathcal{L}_{wpl} + w_3 \times \mathcal{L}_{sem} + w_4 \times \mathcal{L}_{feat} \ ,
\end{equation}
where $w_1,w_2,w_3,w_4$ are weights to balance the loss values.

\subsection{Training Strategies}
\label{sec:training}

Training the model on the unlabeled set, particularly in the early stages, is challenging due to the domain gap between the real and synthetic data. We introduce a strategy to expedite model training by leveraging existing image restoration methods and our proposed VLM-based image assessment.
Additionally, we enhance model performance through iterative updates of pseudo-labels, weather prompts, and descriptions in rounds.

\subsubsection{Pseudo-label initialization.}
In the pseudo-label initialization stage, we gather the initial pseudo-labels by collecting the noisy restoration outcomes from both existing weather-specific and all-in-one image restoration methods. 
Subsequently, we employ the VLM-based image assessment technique to filter out the noisy samples and select the best-restored images as the pseudo-labels to initialize the pseudo-label database. 
To mitigate potential biases from a single vision-language model towards a specific image appearance, we utilize a diverse set of VLMs with varying architectures and parameters as experts for image assessment.

\subsubsection{Iterative update.}
Leveraging expertise from multiple VLMs for image assessment to select pseudo-labels enhances the model learning process.
However, due to computational constraints, it is impractical to consult every VLM during online learning. 
Instead, we divide the overall training into multiple rounds. 
In each round, only one VLM is employed for online image assessment. 
After a round of training, the overall assessment using the set of VLMs for pseudo-labels is conducted, incorporating new predictions from the updated model.
Additionally, we update the weather prompts and augment the descriptions progressively.
Note that the round number is empirically set as four during the training.

\section{Experimental Results}
\label{sec:experiment}

\subsection{Experimental Settings}

\subsubsection{Training \& testing sets.}
We adopt several (pseudo-)synthetic datasets for deraining, dehazing, and desnowing, including Outdoor-Rain~\cite{li2019heavy}, RainDrop~\cite{qian2018attentive}, SPA~\cite{wang2019spatial}, OTS~\cite{li2018benchmarking}, and Snow100K~\cite{liu2018desnownet}.
Meanwhile, we leverage the unlabeled real-world adverse weather images for unsupervised learning.
To achieve this, we utilize the real hazy images in URHI~\cite{li2018benchmarking} (2,318 images) and manually collect real-world rainy and snowy images from the Internet (2,433 and 2,018 images, respectively).
Besides, to train CLIP weather prompts, we employ high-quality DF2K~\cite{timofte2017ntire,lim2017enhanced} images for the \textit{clear} category.
%
We adopt real adverse weather image datasets for qualitative and quantitative evaluation, including RTTS~\cite{li2018benchmarking} with 4,322 haze images, DDN-SIRR~\cite{wei2019semi} and Real3000~\cite{liu2021unpaired} with 2,320 rain images, and Snow100K Realistic~\cite{liu2018desnownet} with 1,329 snow images.
Note that we remove the non-realistic images in Real3000, \eg, comic and movie scenes.

\subsubsection{Implementation details.}
Our semi-supervised learning framework is easily compatible with various image restoration networks.
We opt for MSBDN~\cite{jiang2020multi} as our backbone due to its balanced performance and rapid inference speed in our main study.
Each batch comprises eight labeled and unlabeled images, with a training process spanning 40,000 iterations per round. Image assessment utilizes recent VLMs~\cite{liu2024improved,liu2024llavanext,ye2024mplug,chen2024internvl,sun2024generative}.
Pseudo-labels are initialized through existing weather-specific and all-in-one adverse weather restoration methods.
Empirical values for $w_1, w_2, w_3, w_4$ are set as $0.5, 0.2, 0.05, 0.2$, respectively.
Implementation is based on BasicSR~\cite{basicsr} and training is performed on two NVIDIA A40 GPUs.
%

\subsection{Comparisons with the State-of-the-Art Methods}

\begin{table*}[t]
\centering
\caption{Quantitative comparisons of deraining/dehazing/desnowing on real photos.}
\label{tab:no-ref}
\begin{adjustbox}{width=\hsize}
\begin{tabular}{ccccc}

\toprule
\multicolumn{1}{c|}{\multirow{2}{*}{Method}}                        & \multicolumn{4}{c}{NIMA~\cite{talebi2018nima} $\uparrow$ / MUSIQ~\cite{ke2021musiq} $\uparrow$ / CLIP-IQA~\cite{wang2023exploring} $\uparrow$}                                 \\ \cmidrule{2-5} 
\multicolumn{1}{c|}{}                                               & \multicolumn{1}{c|}{Rain}                      & \multicolumn{1}{c|}{Haze}                      & \multicolumn{1}{c|}{Snow}                      & Overall                     \\ \midrule
\multicolumn{1}{c|}{Restormer~\cite{zamir2022restormer}}            & \multicolumn{1}{c|}{\,5.151 / 54.69 / 0.437\,} & \multicolumn{1}{c|}{\,4.804 / 53.27 / 0.366\,} & \multicolumn{1}{c|}{\,5.020 / 61.18 / 0.510\,} & {\,4.992 / 56.38 / 0.438\,} \\
\multicolumn{1}{c|}{TransWeather~\cite{valanarasu2022transweather}} & \multicolumn{1}{c|}{\,5.068 / 51.06 / 0.358\,} & \multicolumn{1}{c|}{\,4.716 / 46.27 / 0.292\,} & \multicolumn{1}{c|}{\,4.928 / 59.38 / 0.416\,} & {\,4.904 / 52.24 / 0.355\,} \\
\multicolumn{1}{c|}{TKL~\cite{chen2022learning}}                    & \multicolumn{1}{c|}{\,5.099 / 50.96 / 0.392\,} & \multicolumn{1}{c|}{\,4.697 / 48.21 / 0.318\,} & \multicolumn{1}{c|}{\,4.905 / 59.24 / 0.428\,} & {\,4.900 / 52.80 / 0.379\,} \\
\multicolumn{1}{c|}{WeatherDiff~\cite{ozdenizci2023restoring}}      & \multicolumn{1}{c|}{\,5.054 / 51.82 / 0.395\,} & \multicolumn{1}{c|}{\,4.616 / 47.70 / 0.326\,} & \multicolumn{1}{c|}{\,4.917 / 60.52 / 0.466\,} & {\,4.862 / 53.35 / 0.396\,} \\
\multicolumn{1}{c|}{WGWS-Net~\cite{zhu2023learning}}                & \multicolumn{1}{c|}{\,5.035 / 51.46 / 0.389\,} & \multicolumn{1}{c|}{\,4.815 / 45.76 / 0.310\,} & \multicolumn{1}{c|}{\,4.779 / 57.95 / 0.395\,} & {\,4.876 / 51.72 / 0.365\,} \\
\multicolumn{1}{c|}{MWDT~\cite{patil2023multi}}                     & \multicolumn{1}{c|}{\,5.104 / 52.47 / 0.377\,} & \multicolumn{1}{c|}{\,4.741 / 51.23 / 0.315\,} & \multicolumn{1}{c|}{\,5.034 / 60.16 / 0.407\,} & {\,4.960 / 54.62 / 0.366\,} \\
\multicolumn{1}{c|}{PromptIR~\cite{potlapalli2023promptir}}         & \multicolumn{1}{c|}{\,5.174 / 53.48 / 0.439\,} & \multicolumn{1}{c|}{\,4.823 / 53.88 / \textbf{0.372}\,} & \multicolumn{1}{c|}{\,5.032 / 60.86 / 0.517\,} & {\,5.009 / 56.07 / 0.443\,} \\
\multicolumn{1}{c|}{DA-CLIP~\cite{luo2024controlling}}              & \multicolumn{1}{c|}{\,5.168 / 52.98 / 0.412\,} & \multicolumn{1}{c|}{\,4.851 / 53.23 / 0.325\,} & \multicolumn{1}{c|}{\,5.012 / 60.57 / 0.499\,} & {\,5.010 / 55.59 / 0.412\,} \\ \midrule
\multicolumn{1}{c|}{Our method}                                     & \multicolumn{1}{c|}{\,\textbf{5.291} / \textbf{59.80} / \textbf{0.477}\,} & \multicolumn{1}{c|}{\,\textbf{4.906} / \textbf{56.09} / 0.371\,} & \multicolumn{1}{c|}{\,\textbf{5.057} / \textbf{62.12} / \textbf{0.519}\,} & {\,\textbf{5.084} / \textbf{59.34} / \textbf{0.456}\,} \\
\bottomrule
                                                                    &                                                &                                                &                                                &                             \\
\toprule
\multicolumn{1}{c|}{\multirow{2}{*}{Method}}                        & \multicolumn{4}{c}{LIQE~\cite{zhang2023blind} / Q-Align~\cite{wu2024qalign} $\uparrow$ / VLM-Vis $\uparrow$}                                                                   \\ \cmidrule{2-5} 
\multicolumn{1}{c|}{}                                               & \multicolumn{1}{c|}{Rain}                      & \multicolumn{1}{c|}{Haze}                      & \multicolumn{1}{c|}{Snow}                      & Overall                     \\ \midrule
\multicolumn{1}{c|}{Restormer~\cite{zamir2022restormer}}            & \multicolumn{1}{c|}{\,2.277 / 3.795 / 0.417\,} & \multicolumn{1}{c|}{\,1.918 / 3.068 / 0.218\,} & \multicolumn{1}{c|}{\,3.172 / 3.646 / 0.395\,} & {\,2.456 / 3.503 / 0.343\,} \\
\multicolumn{1}{c|}{TransWeather~\cite{valanarasu2022transweather}} & \multicolumn{1}{c|}{\,1.924 / 3.545 / 0.402\,} & \multicolumn{1}{c|}{\,1.502 / 2.809 / 0.223\,} & \multicolumn{1}{c|}{\,2.770 / 3.537 / 0.384\,} & {\,2.065 / 3.297 / 0.336\,} \\
\multicolumn{1}{c|}{TKL~\cite{chen2022learning}}                    & \multicolumn{1}{c|}{\,2.028 / 3.588 / 0.406\,} & \multicolumn{1}{c|}{\,1.590 / 2.908 / 0.238\,} & \multicolumn{1}{c|}{\,2.830 / 3.557 / 0.393\,} & {\,2.149 / 3.351 / 0.346\,} \\
\multicolumn{1}{c|}{WeatherDiff~\cite{ozdenizci2023restoring}}      & \multicolumn{1}{c|}{\,2.050 / 3.640 / 0.411\,} & \multicolumn{1}{c|}{\,1.520 / 2.843 / 0.217\,} & \multicolumn{1}{c|}{\,2.950 / 3.573 / 0.397\,} & {\,2.173 / 3.352 / 0.342\,} \\
\multicolumn{1}{c|}{WGWS-Net~\cite{zhu2023learning}}                & \multicolumn{1}{c|}{\,1.965 / 3.592 / 0.411\,} & \multicolumn{1}{c|}{\,1.506 / 2.915 / 0.238\,} & \multicolumn{1}{c|}{\,2.619 / 3.490 / 0.383\,} & {\,2.030 / 3.332 / 0.344\,} \\
\multicolumn{1}{c|}{MWDT~\cite{patil2023multi}}                     & \multicolumn{1}{c|}{\,2.068 / 3.548 / 0.426\,} & \multicolumn{1}{c|}{\,1.720 / 2.861 / 0.273\,} & \multicolumn{1}{c|}{\,2.903 / 3.569 / 0.412\,} & {\,2.230 / 3.326 / 0.370\,} \\
\multicolumn{1}{c|}{PromptIR~\cite{potlapalli2023promptir}}         & \multicolumn{1}{c|}{\,2.250 / 3.770 / 0.419\,} & \multicolumn{1}{c|}{\,1.941 / 3.093 / 0.226\,} & \multicolumn{1}{c|}{\,3.121 / 3.609 / 0.384\,} & {\,2.437 / 3.491 / 0.343\,} \\
\multicolumn{1}{c|}{DA-CLIP~\cite{luo2024controlling}}              & \multicolumn{1}{c|}{\,2.250 / 3.732 / 0.412\,} & \multicolumn{1}{c|}{\,2.014 / 3.071 / 0.230\,} & \multicolumn{1}{c|}{\,3.050 / 3.637 / 0.395\,} & {\,2.438 / 3.480 / 0.346\,} \\ \midrule
\multicolumn{1}{c|}{Our method}                                     & \multicolumn{1}{c|}{\,\textbf{2.563} / \textbf{3.843} / \textbf{0.440}\,} & \multicolumn{1}{c|}{\,\textbf{2.064} / \textbf{3.176} / \textbf{0.289}\,} & \multicolumn{1}{c|}{\,\textbf{3.293} / \textbf{3.702} / \textbf{0.431}\,} & {\,\textbf{2.640} / \textbf{3.574} / \textbf{0.387}\,} \\
\bottomrule

\end{tabular}
\end{adjustbox}
\end{table*}

We benchmark our method against several state-of-the-art general and all-in-one adverse weather image restoration approaches. Our comparisons encompass recent works including Restormer~\cite{zamir2022restormer}, TransWeather~\cite{valanarasu2022transweather}, TKL~\cite{chen2022learning}, WeatherDiff~\cite{ozdenizci2023restoring}, WGWS-Net~\cite{zhu2023learning}, MWDT~\cite{patil2023multi}, PromptIR~\cite{potlapalli2023promptir}, and DA-CLIP~\cite{luo2024controlling}. 
%
We compare with the best-performing models from either retrained versions using our paired data or officially released checkpoints for fairness.

\subsubsection{Quantitative comparison.}
Note that there is no ground-truth clear image for the real adverse weather images.
Therefore, we adopt several no-reference metrics for the quantitative assessment.
Specifically, we use recent blind image quality evaluation metrics, including NIMA~\cite{talebi2018nima}, MUSIQ~\cite{ke2021musiq}, CLIP-IQA~\cite{wang2023exploring}, LIQE~\cite{zhang2023blind}, and Q-Align~\cite{wu2024qalign}.
We also utilize the proposed VLM-based image visibility assessment method and report the normalized scores VLM-Vis.
In detail, VLM-Vis is computed over VLM experts, standardized by the minimum and maximum statistics across the dataset for each respective VLM.
The quantitative comparisons are reported in \Cref{tab:no-ref}.
Our proposed method is ranked first for all image quality assessment metrics on average and in almost all weather conditions.
These values indicate the superior restoration quality of the images.
Moreover, our method achieves the best VLM-Vis across different weather conditions.
These results demonstrate the advantages of our method on real data against existing advanced adverse weather image restoration methods, which focus mainly on synthetic data evaluation.
%


\subsubsection{Qualitative comparison.}

\begin{figure*}[t]
    \centering
    \captionsetup[subfigure]{labelformat=empty,justification=centering}
    \begin{subfigure}{0.135\hsize}
        \includegraphics[width=\hsize]{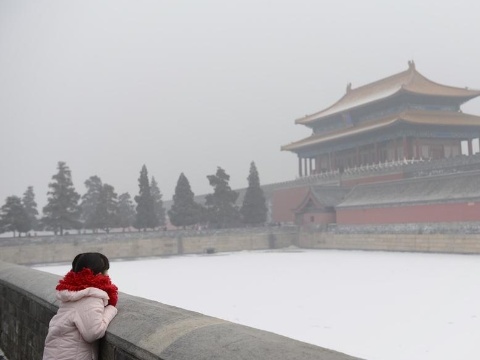}
    \end{subfigure}
    \begin{subfigure}{0.135\hsize}
        \includegraphics[width=\hsize]{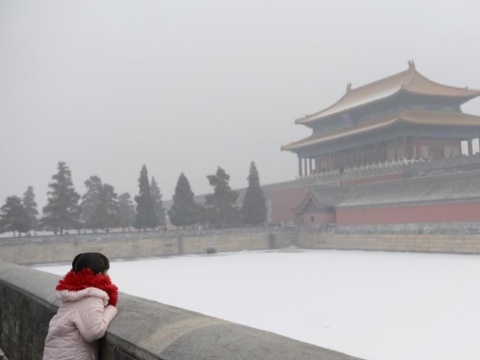}
    \end{subfigure}
    \begin{subfigure}{0.135\hsize}
        \includegraphics[width=\hsize]{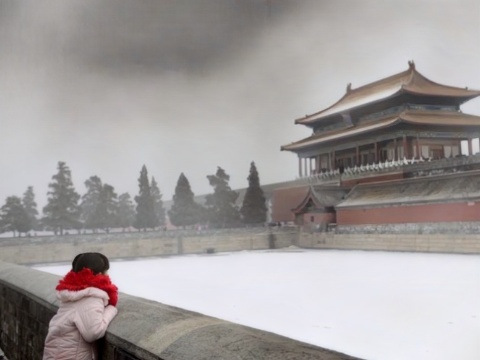}
    \end{subfigure}
    \begin{subfigure}{0.135\hsize}
        \includegraphics[width=\hsize]{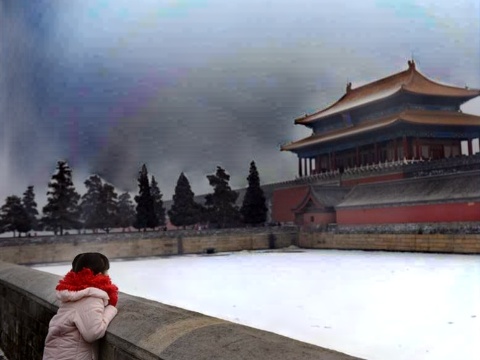}
    \end{subfigure}
    \begin{subfigure}{0.135\hsize}
        \includegraphics[width=\hsize]{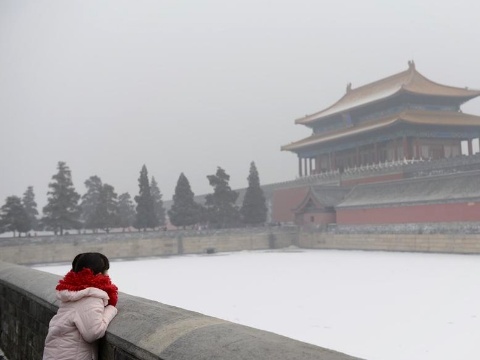}
    \end{subfigure}
    \begin{subfigure}{0.135\hsize}
        \includegraphics[width=\hsize]{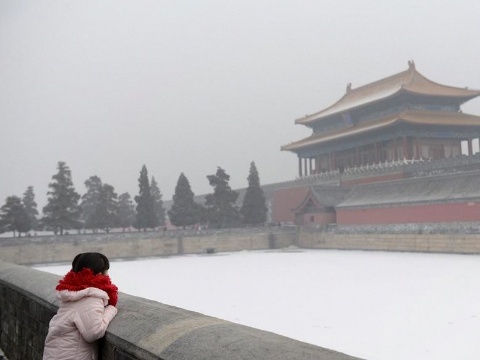}
    \end{subfigure}
    \begin{subfigure}{0.135\hsize}
        \includegraphics[width=\hsize]{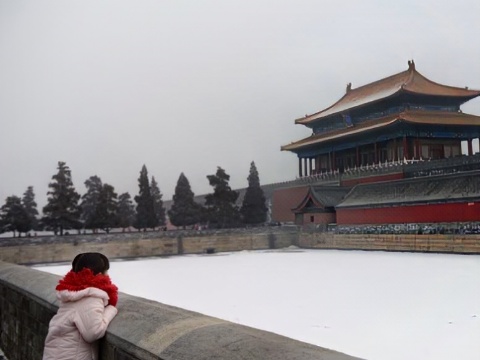}
    \end{subfigure}
    \\
    \begin{subfigure}{0.135\hsize}
        \includegraphics[width=\hsize]{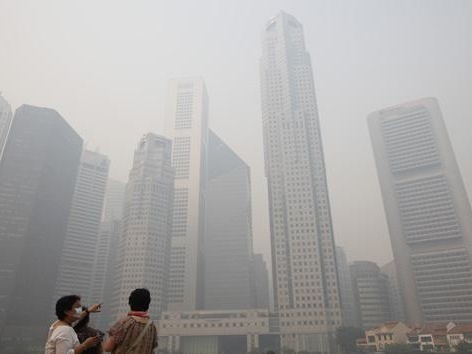}
    \end{subfigure}
    \begin{subfigure}{0.135\hsize}
        \includegraphics[width=\hsize]{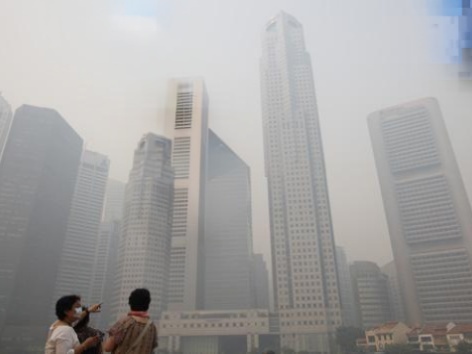}
    \end{subfigure}
    \begin{subfigure}{0.135\hsize}
        \includegraphics[width=\hsize]{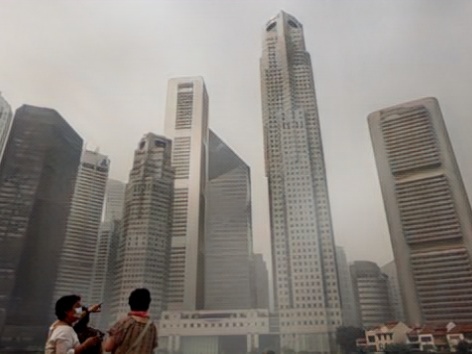}
    \end{subfigure}
    \begin{subfigure}{0.135\hsize}
        \includegraphics[width=\hsize]{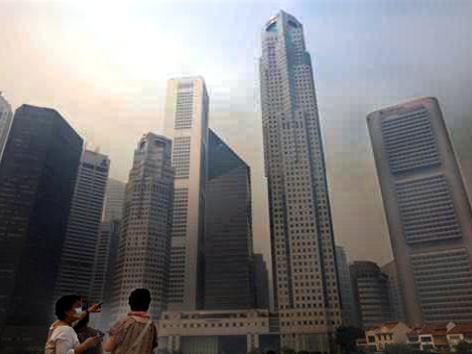}
    \end{subfigure}
    \begin{subfigure}{0.135\hsize}
        \includegraphics[width=\hsize]{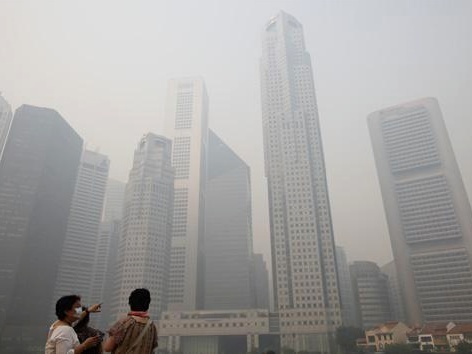}
    \end{subfigure}
    \begin{subfigure}{0.135\hsize}
        \includegraphics[width=\hsize]{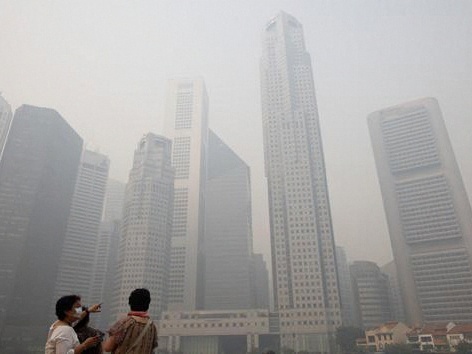}
    \end{subfigure}
    \begin{subfigure}{0.135\hsize}
        \includegraphics[width=\hsize]{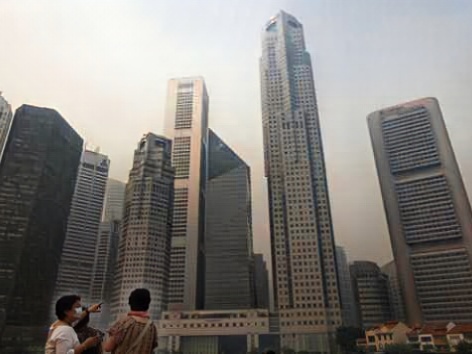}
    \end{subfigure}
    \\
    %
    \begin{subfigure}{0.135\hsize}
        \includegraphics[width=\hsize]{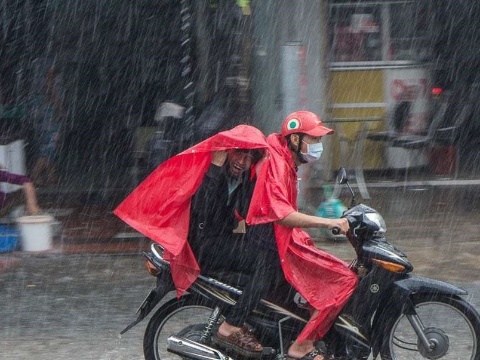}
    \end{subfigure}
    \begin{subfigure}{0.135\hsize}
        \includegraphics[width=\hsize]{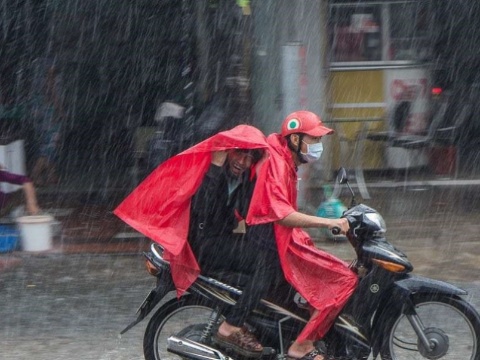}
    \end{subfigure}
    \begin{subfigure}{0.135\hsize}
        \includegraphics[width=\hsize]{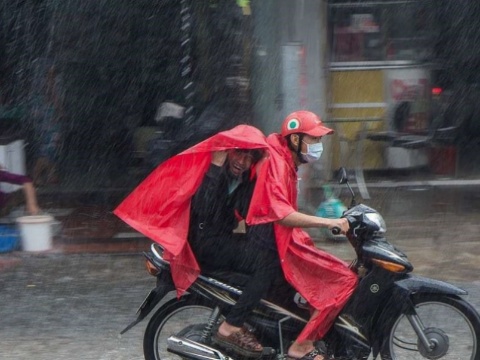}
    \end{subfigure}
    \begin{subfigure}{0.135\hsize}
        \includegraphics[width=\hsize]{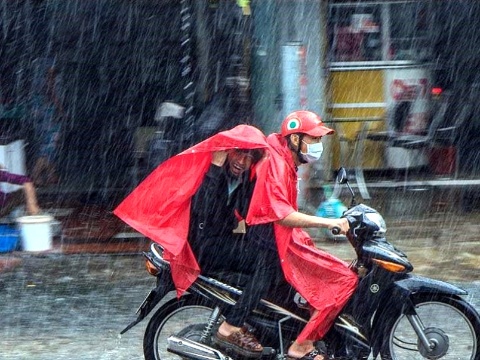}
    \end{subfigure}
    \begin{subfigure}{0.135\hsize}
        \includegraphics[width=\hsize]{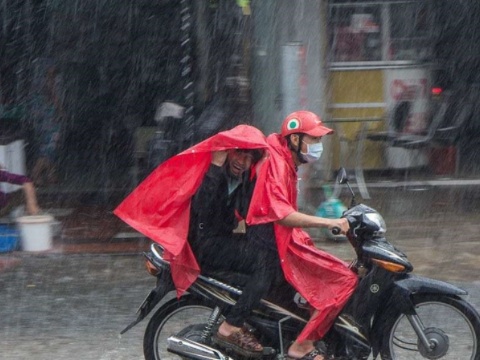}
    \end{subfigure}
    \begin{subfigure}{0.135\hsize}
        \includegraphics[width=\hsize]{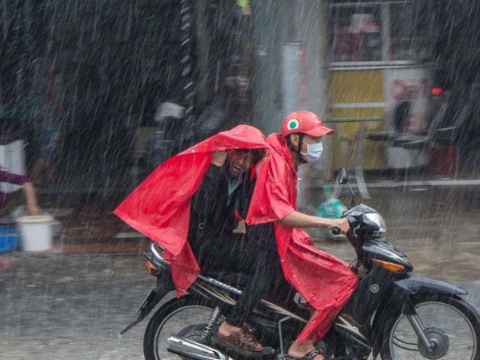}
    \end{subfigure}
    \begin{subfigure}{0.135\hsize}
        \includegraphics[width=\hsize]{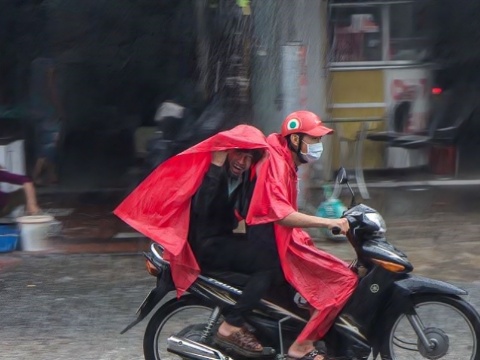}
    \end{subfigure}
    \\
    \begin{subfigure}{0.135\hsize}
        \includegraphics[width=\hsize]{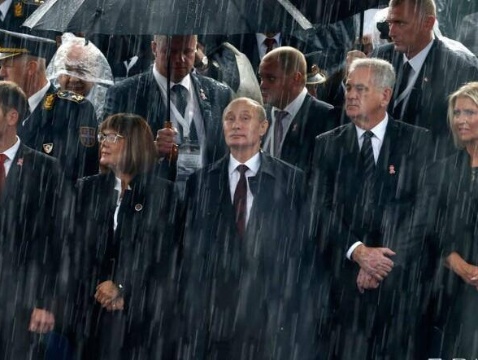}
    \end{subfigure}
    \begin{subfigure}{0.135\hsize}
        \includegraphics[width=\hsize]{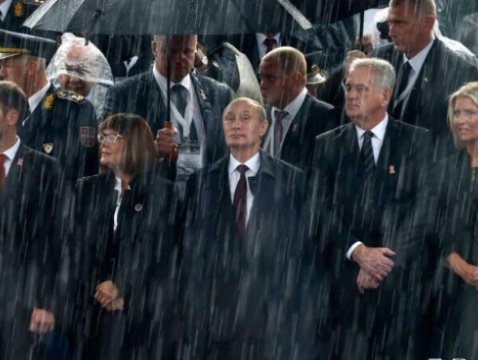}
    \end{subfigure}
    \begin{subfigure}{0.135\hsize}
        \includegraphics[width=\hsize]{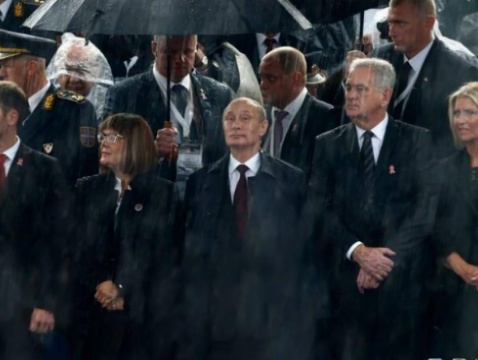}
    \end{subfigure}
    \begin{subfigure}{0.135\hsize}
        \includegraphics[width=\hsize]{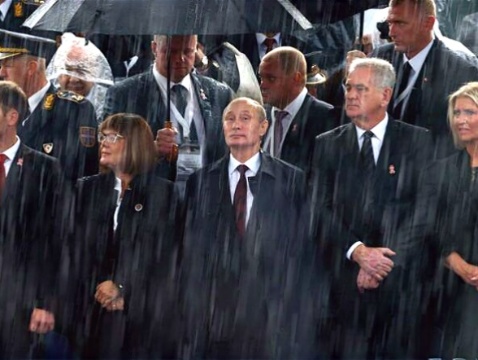}
    \end{subfigure}
    \begin{subfigure}{0.135\hsize}
        \includegraphics[width=\hsize]{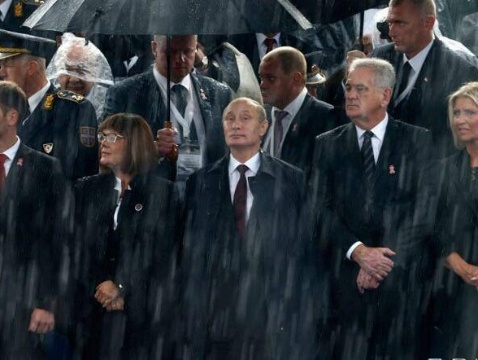}
    \end{subfigure}
    \begin{subfigure}{0.135\hsize}
        \includegraphics[width=\hsize]{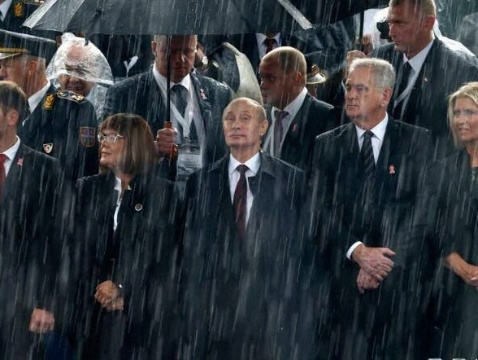}
    \end{subfigure}
    \begin{subfigure}{0.135\hsize}
        \includegraphics[width=\hsize]{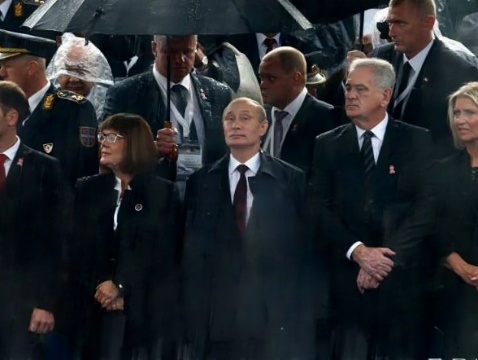}
    \end{subfigure}
    \\
    \begin{subfigure}{0.135\hsize}
        \includegraphics[width=\hsize]{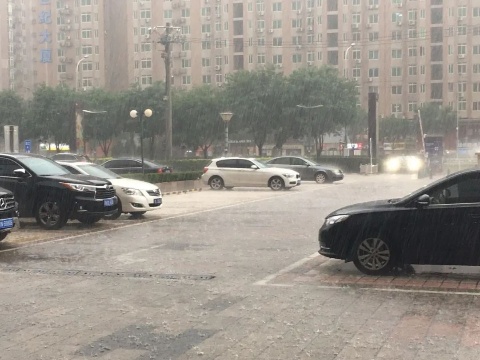}
    \end{subfigure}
    \begin{subfigure}{0.135\hsize}
        \includegraphics[width=\hsize]{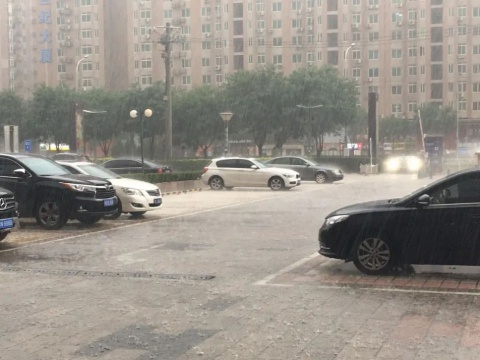}
    \end{subfigure}
    \begin{subfigure}{0.135\hsize}
        \includegraphics[width=\hsize]{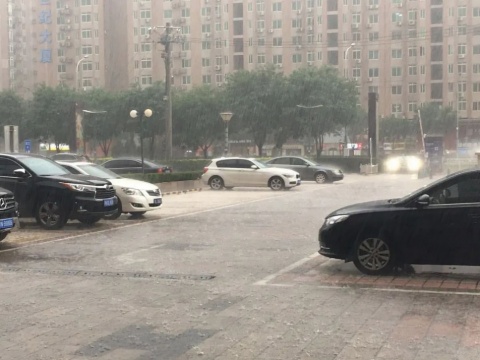}
    \end{subfigure}
    \begin{subfigure}{0.135\hsize}
        \includegraphics[width=\hsize]{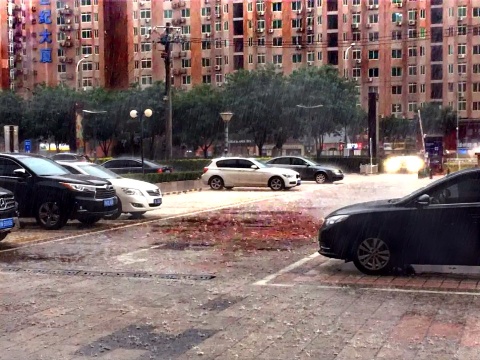}
    \end{subfigure}
    \begin{subfigure}{0.135\hsize}
        \includegraphics[width=\hsize]{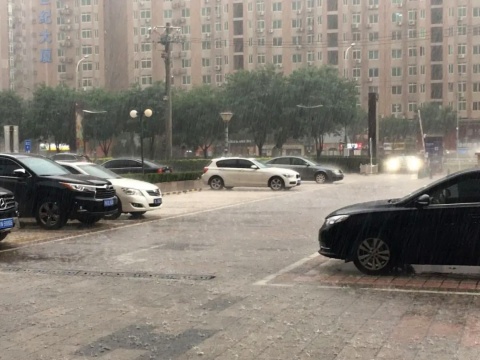}
    \end{subfigure}
    \begin{subfigure}{0.135\hsize}
        \includegraphics[width=\hsize]{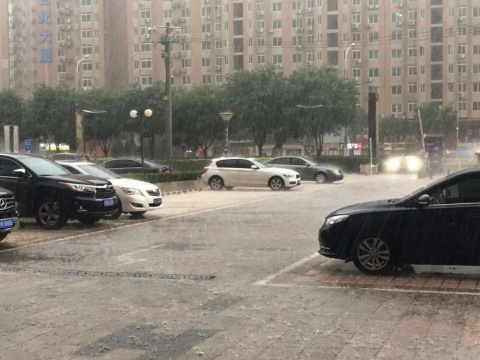}
    \end{subfigure}
    \begin{subfigure}{0.135\hsize}
        \includegraphics[width=\hsize]{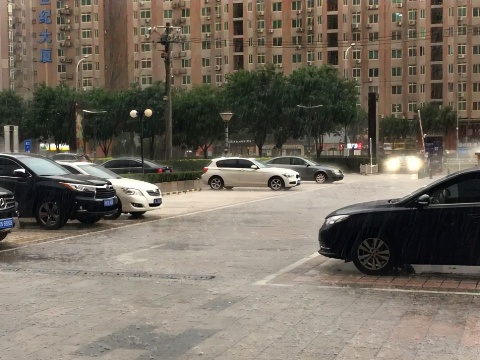}
    \end{subfigure}
    \\
    \begin{subfigure}{0.135\hsize}
        \includegraphics[width=\hsize]{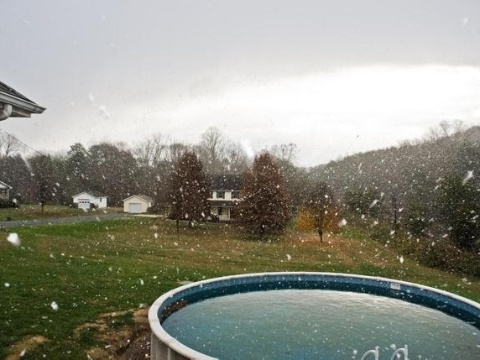}
    \end{subfigure}
    \begin{subfigure}{0.135\hsize}
        \includegraphics[width=\hsize]{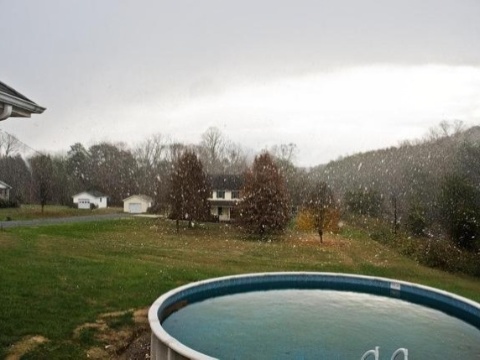}
    \end{subfigure}
    \begin{subfigure}{0.135\hsize}
        \includegraphics[width=\hsize]{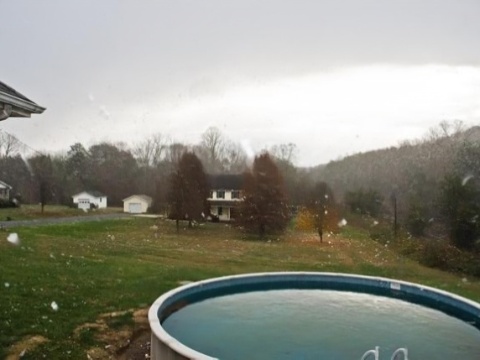}
    \end{subfigure}
    \begin{subfigure}{0.135\hsize}
        \includegraphics[width=\hsize]{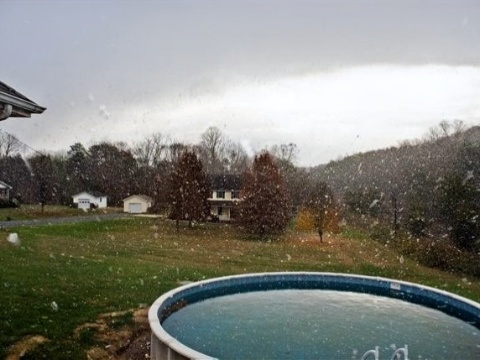}
    \end{subfigure}
    \begin{subfigure}{0.135\hsize}
        \includegraphics[width=\hsize]{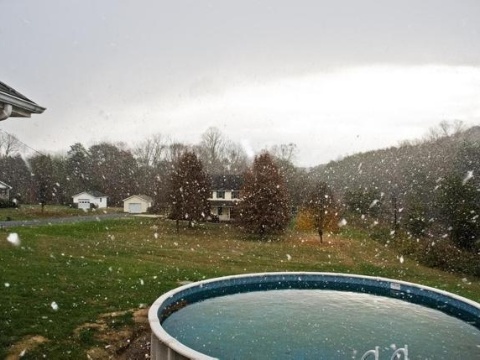}
    \end{subfigure}
    \begin{subfigure}{0.135\hsize}
        \includegraphics[width=\hsize]{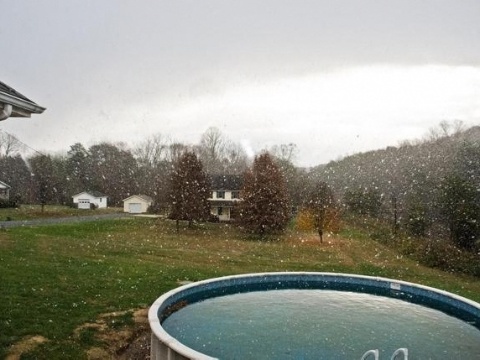}
    \end{subfigure}
    \begin{subfigure}{0.135\hsize}
        \includegraphics[width=\hsize]{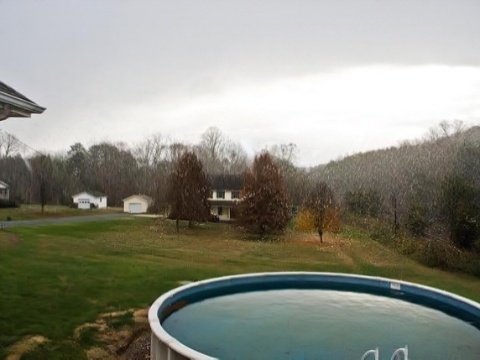}
    \end{subfigure}
    \\
    \begin{subfigure}{0.135\hsize}
        \includegraphics[width=\hsize]{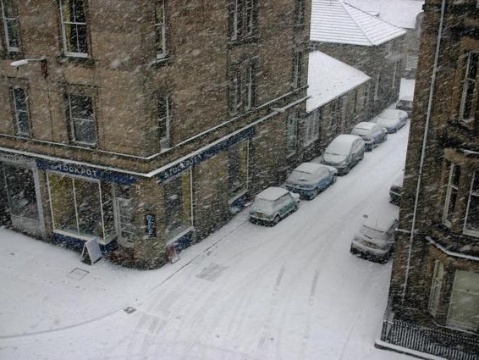}
        \caption{Input}
    \end{subfigure}
    \begin{subfigure}{0.135\hsize}
        \includegraphics[width=\hsize]{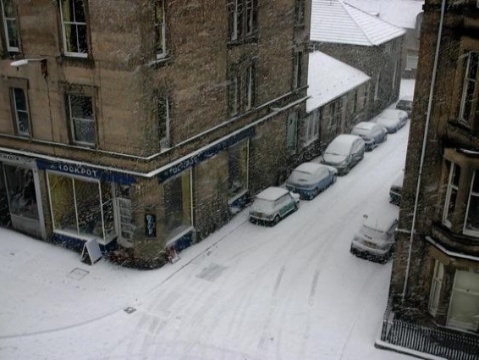}
        \caption{WeatherDiff}
    \end{subfigure}
    \begin{subfigure}{0.135\hsize}
        \includegraphics[width=\hsize]{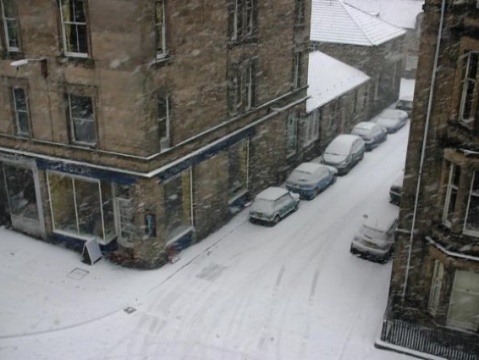}
        \caption{WGWS-Net}
    \end{subfigure}
    \begin{subfigure}{0.135\hsize}
        \includegraphics[width=\hsize]{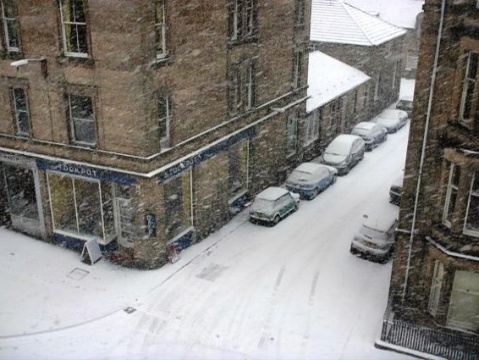}
        \caption{MWDT}
    \end{subfigure}
    \begin{subfigure}{0.135\hsize}
        \includegraphics[width=\hsize]{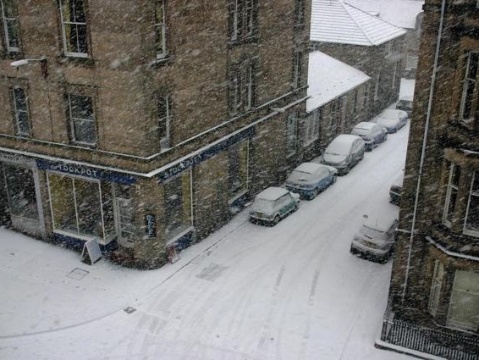}
        \caption{PromptIR}
    \end{subfigure}
    \begin{subfigure}{0.135\hsize}
        \includegraphics[width=\hsize]{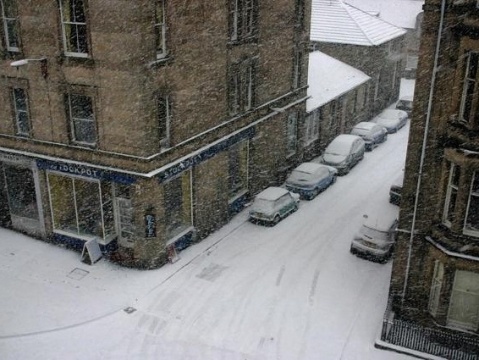}
        \caption{DA-CLIP}
    \end{subfigure}
    \begin{subfigure}{0.135\hsize}
        \includegraphics[width=\hsize]{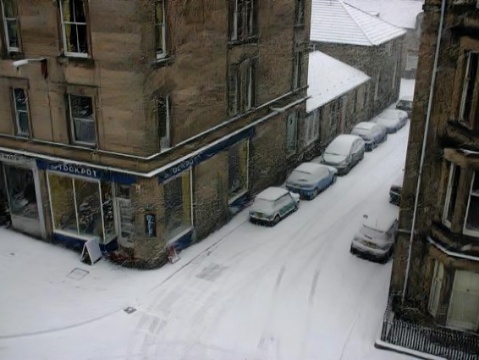}
        \caption{Ours}
    \end{subfigure}
    \\
    \caption{Visual comparisons on real-world images~\cite{li2018benchmarking,wei2019semi,liu2021unpaired,liu2018desnownet}.
    }
    \label{fig:vis_real}
\end{figure*}

Our qualitative assessment is conducted on real-world evaluation datasets~\cite{li2018benchmarking,wei2019semi,liu2021unpaired,liu2018desnownet} and the visual outcomes are presented in \cref{fig:vis_real}.
We can observe that the compared methods are less effective in dealing with real-world adverse weather images and are limited in removing rain, haze, and snow artifacts.
It is noted that MWDT~\cite{patil2023multi} mitigates the haze effect but introduces severe color distortion. 
In comparison, our method exhibits superior visually perceptual quality, enhancing clarity and contrast, while minimizing rain, haze, and snow artifacts. Notably, our approach effectively eliminates haze in rain and snow scenarios, significantly improving image visibility.

\subsubsection{User study.}

\begin{figure}[bp]
    \centering
    \includegraphics[width=0.72\hsize]{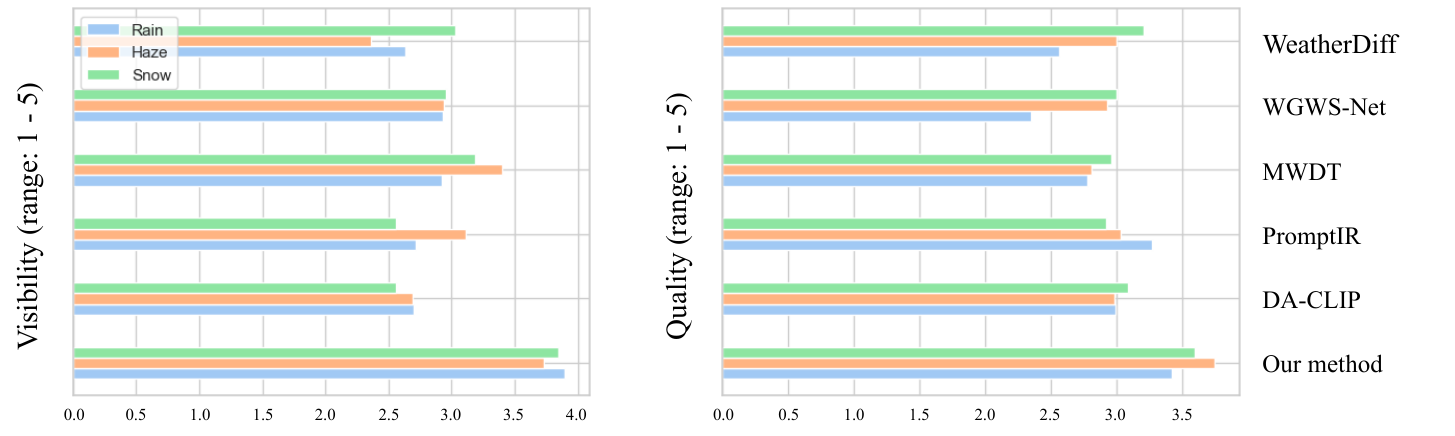}
    \caption{User study on visibility and quality of image restoration.}
    \label{fig:userstudy}
\end{figure}

We conducted a user study to evaluate the visual quality.
For each weather scenario, ten real-world images are chosen.
32 participants were invited for the evaluation.
Two factors are considered, \ie, image visibility and quality, regarding the extent to which the weather-related artifacts are removed and the restored image is kept real.
As observed in \cref{fig:userstudy}, MWDT obtains high image visibility scores, which aligns with our VLM-Vis metric.
Overall, our method exhibits a clear advantage in visibility and quality across weather conditions.

\subsection{Ablation Studies}


\subsubsection{Effectiveness of semi-supervised learning framework.}

\begin{figure*}[t]
    \centering
    \captionsetup[subfigure]{labelformat=empty,justification=centering}
    \begin{subfigure}{0.115\hsize}
        \includegraphics[width=\hsize]{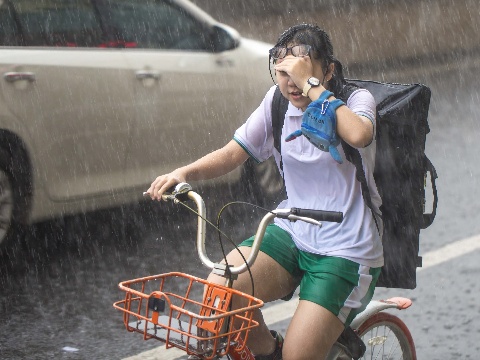}
    \end{subfigure}
    \begin{subfigure}{0.115\hsize}
        \includegraphics[width=\hsize]{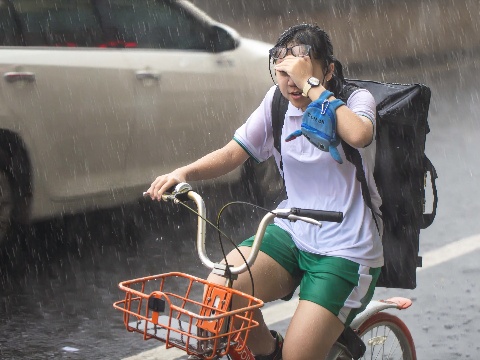}
    \end{subfigure}
    \begin{subfigure}{0.115\hsize}
        \includegraphics[width=\hsize]{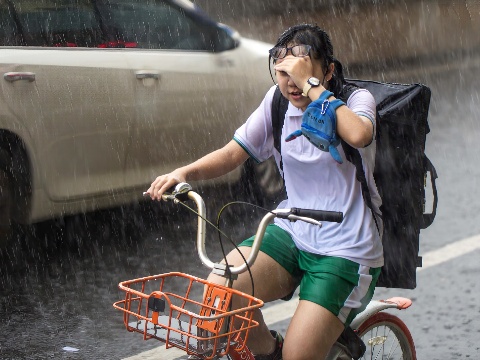}
    \end{subfigure}
    \begin{subfigure}{0.115\hsize}
        \includegraphics[width=\hsize]{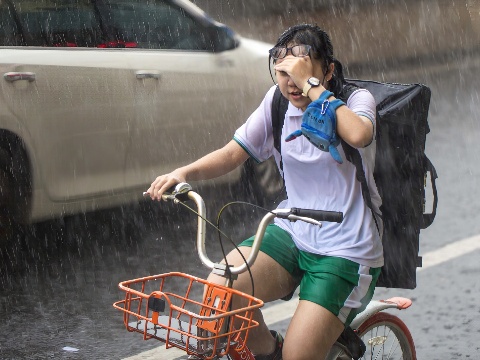}
    \end{subfigure}
    \begin{subfigure}{0.115\hsize}
        \includegraphics[width=\hsize]{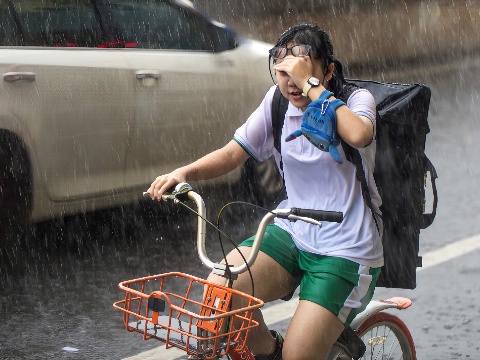}
    \end{subfigure}
    \begin{subfigure}{0.115\hsize}
        \includegraphics[width=\hsize]{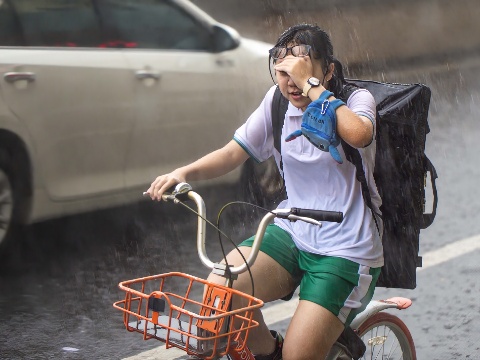}
    \end{subfigure}
    \begin{subfigure}{0.115\hsize}
        \includegraphics[width=\hsize]{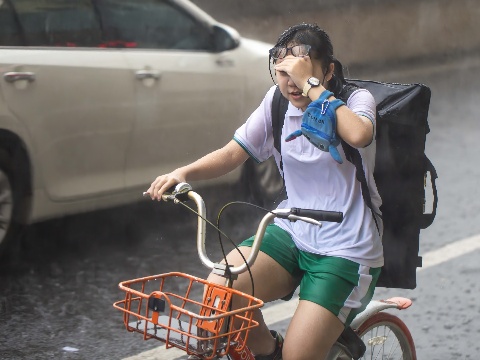}
    \end{subfigure}
    \begin{subfigure}{0.115\hsize}
        \includegraphics[width=\hsize]{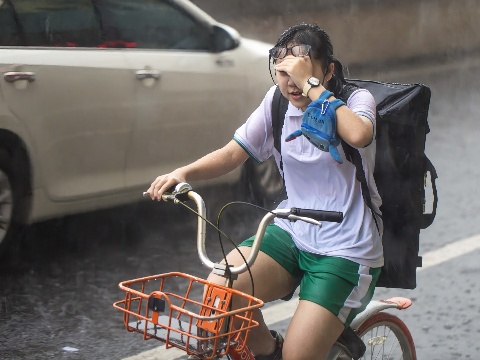}
    \end{subfigure}
    \\
    \begin{subfigure}{0.115\hsize}
        \includegraphics[width=\hsize]{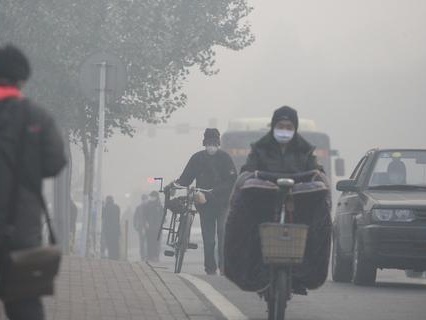}
    \end{subfigure}
    \begin{subfigure}{0.115\hsize}
        \includegraphics[width=\hsize]{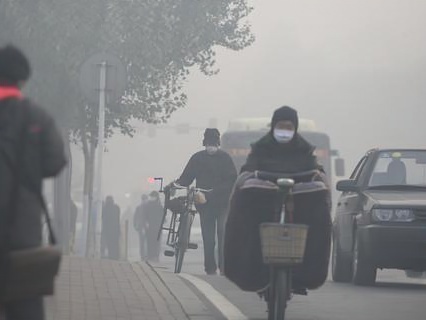}
    \end{subfigure}
    \begin{subfigure}{0.115\hsize}
        \includegraphics[width=\hsize]{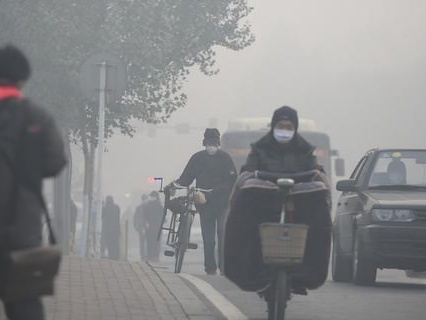}
    \end{subfigure}
    \begin{subfigure}{0.115\hsize}
        \includegraphics[width=\hsize]{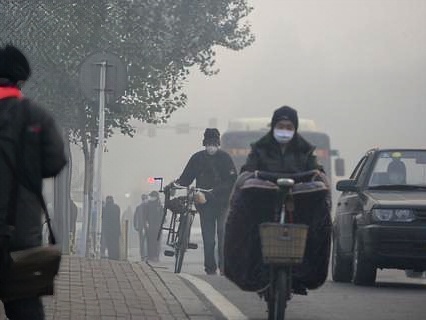}
    \end{subfigure}
    \begin{subfigure}{0.115\hsize}
        \includegraphics[width=\hsize]{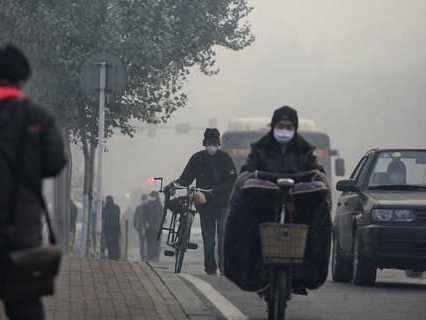}
    \end{subfigure}
    \begin{subfigure}{0.115\hsize}
        \includegraphics[width=\hsize]{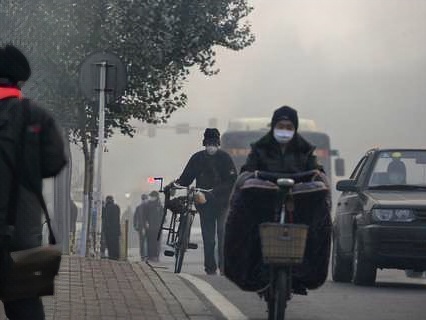}
    \end{subfigure}
    \begin{subfigure}{0.115\hsize}
        \includegraphics[width=\hsize]{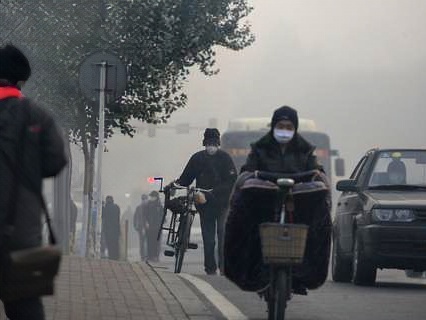}
    \end{subfigure}
    \begin{subfigure}{0.115\hsize}
        \includegraphics[width=\hsize]{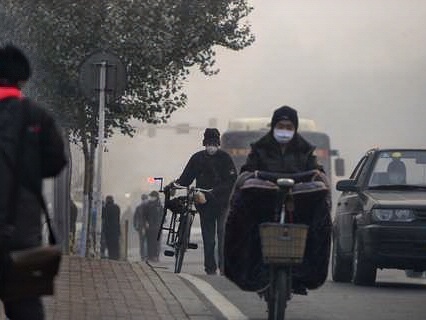}
    \end{subfigure}
    \\
    \begin{subfigure}{0.115\hsize}
        \includegraphics[width=\hsize]{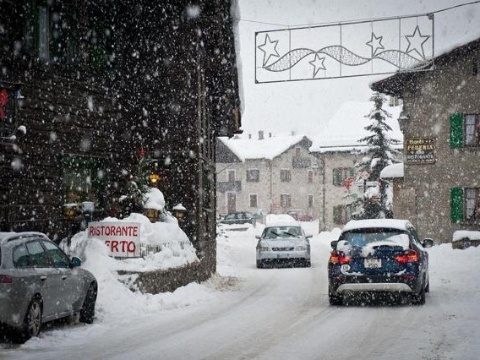}
        \caption{Input}
    \end{subfigure}
    \begin{subfigure}{0.115\hsize}
        \includegraphics[width=\hsize]{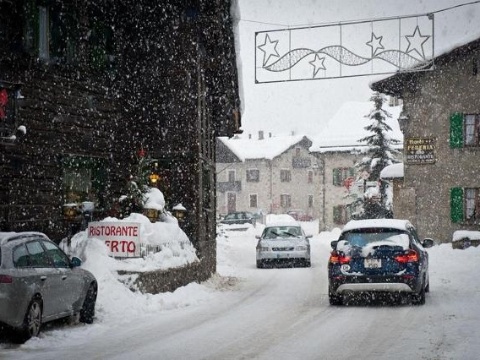}
        \caption{$\mathcal{L}_{sup}$}
    \end{subfigure}
    \begin{subfigure}{0.115\hsize}
        \includegraphics[width=\hsize]{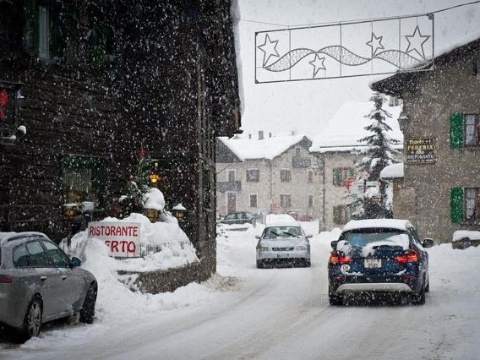}
        \caption{$+ \ \mathcal{L}_{ps}$}
    \end{subfigure}
    \begin{subfigure}{0.115\hsize}
        \includegraphics[width=\hsize]{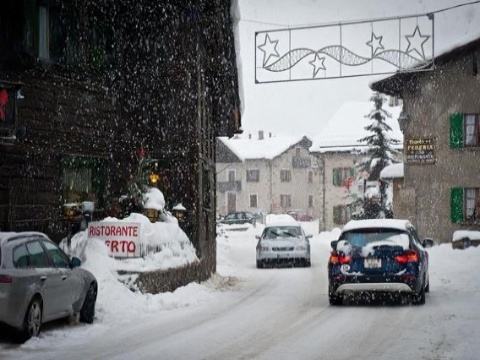}
        \caption{$+ \ r^{vlm}$}
    \end{subfigure}
    \begin{subfigure}{0.115\hsize}
        \includegraphics[width=\hsize]{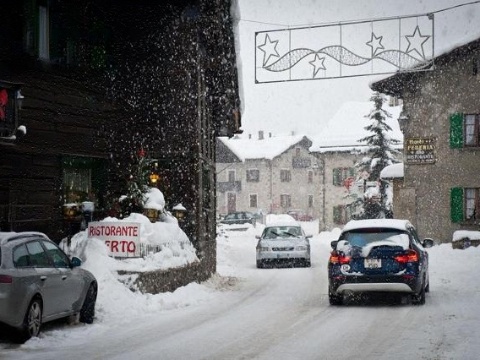}
        \caption{$+$ \textit{init}}
    \end{subfigure}
    \begin{subfigure}{0.115\hsize}
        \includegraphics[width=\hsize]{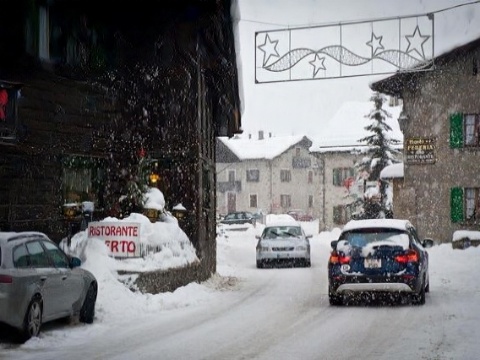}
        \caption{$+ \ \mathcal{L}_{wpl}$}
    \end{subfigure}
    \begin{subfigure}{0.115\hsize}
        \includegraphics[width=\hsize]{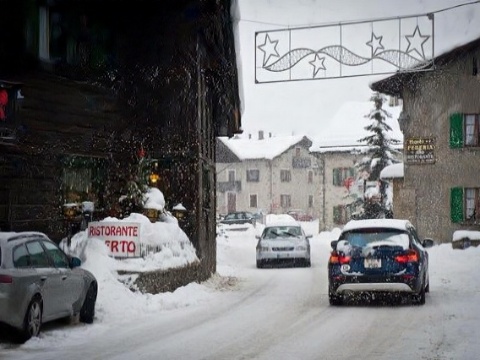}
        \caption{$+ \ \mathcal{L}_{sem}$}
    \end{subfigure}
    \begin{subfigure}{0.115\hsize}
        \includegraphics[width=\hsize]{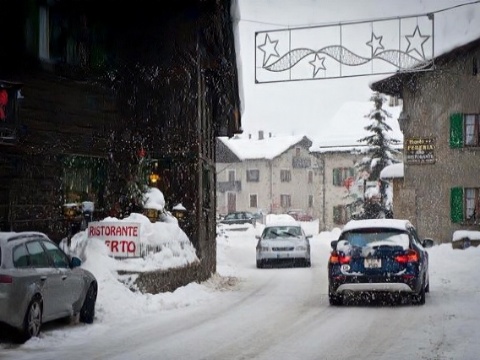}
        \caption{Ours}
    \end{subfigure}
    \\
    \caption{Ablation studies of the proposed semi-supervised learning framework.
    }
    \label{fig:vis_ablation-main}
\end{figure*}

\begin{table}[tb]
\centering
\caption{Ablation analysis of each component design.}
\label{tab:ablation-main}
\scriptsize
\begin{tabular}{ccccccccccc}
\toprule
$\mathcal{L}_{sup}$ & $\mathcal{L}_{ps}$ & $r^{vlm}$ & \textit{init} & $\mathcal{L}_{wpl}$ & $\mathcal{L}_{sem}$ & \textit{iter} &  & MUSIQ $\uparrow$ & CLIP-IQA $\uparrow$ & VLM-Vis $\uparrow$ \\
\midrule
\ding{51}           &                    &           &               &                     &                     &               &  & 53.41            & 0.388               & 0.343              \\
\ding{51}           & \ding{51}          &           &               &                     &                     &               &  & 54.08            & 0.396               & 0.354              \\
\ding{51}           & \ding{51}          & \ding{51} &               &                     &                     &               &  & 56.68            & 0.429               & 0.366              \\
\ding{51}           & \ding{51}          & \ding{51} & \ding{51}     &                     &                     &               &  & 57.34            & 0.425               & 0.370              \\
\ding{51}           & \ding{51}          & \ding{51} & \ding{51}     & \ding{51}           &                     &               &  & 58.13            & 0.437               & 0.376              \\
\ding{51}           & \ding{51}          & \ding{51} & \ding{51}     & \ding{51}           & \ding{51}           &               &  & 58.91            & 0.445               & 0.381              \\
\ding{51}           & \ding{51}          & \ding{51} & \ding{51}     & \ding{51}           & \ding{51}           & \ding{51}     &  & \textbf{59.34}   & \textbf{0.456}      & \textbf{0.387}     \\
\bottomrule
\end{tabular}
\end{table}

We start with the baseline model, trained exclusively through supervised learning ($\mathcal{L}_{sup}$) on labeled synthetic data. Subsequently, we employ the naive mean-teacher~\cite{tarvainen2017mean}, a semi-supervised learning method, to explore unlabeled real data and utilize predictions from the teacher network as pseudo-labels ($\mathcal{L}_{ps}$).
We investigate the effectiveness of the proposed VLM-based components and training strategies, including:
(1) Incorporating VLM-based image assessment $r^{vlm}$ for updating pseudo-labels,
(2) Pseudo-label initialization (\textit{init}),
(3) Weather prompt learning ($\mathcal{L}_{wpl}$),
(4) Semantics regularization ($\mathcal{L}_{sem}$), and
(5) Iterative update (\textit{iter}).


Quantitative outcomes with overall performance across different weather conditions are presented in \Cref{tab:ablation-main}, while visual comparisons are depicted in \cref{fig:vis_ablation-main}.
It is evident that the baseline, trained solely on synthetic data using a straightforward semi-supervised learning approach, struggles to effectively address real rain, haze, and snow artifacts.
In contrast, our proposed VLM-based image assessment progressively refines the selection of superior pseudo-labels, emphasizing higher clearness and resulting in predictions with improved visibility. This effect is further amplified with the incorporation of the pseudo-label initialization strategy.
Moreover, the proposed weather prompt learning and description-assisted semantic enhancement largely improve the restoration performance.
This is evidenced by the boosted image quality, the visibility metric scores, and the visual quality with reduced weather-related artifacts (\cref{fig:vis_ablation-main}).
Lastly, our iterative training strategy further enhances the overall quantitative and qualitative outcomes.

\subsubsection{Impact of VLM-based image assessment.}

\begin{figure*}[t]
    \centering
    \captionsetup[subfigure]{labelformat=empty,justification=centering}
    \begin{subfigure}{0.12\hsize}
        \includegraphics[width=\hsize]{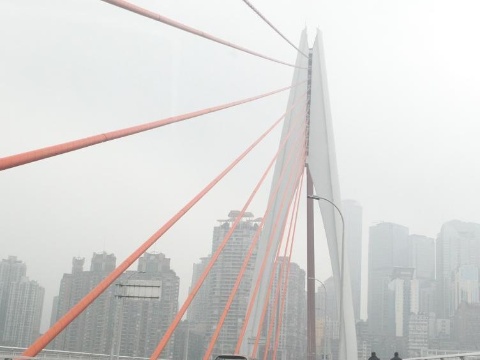}
    \end{subfigure}
    \begin{subfigure}{0.12\hsize}
        \includegraphics[width=\hsize]{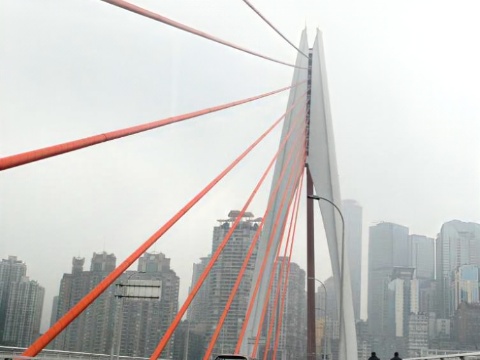}
    \end{subfigure}
    \begin{subfigure}{0.12\hsize}
        \includegraphics[width=\hsize]{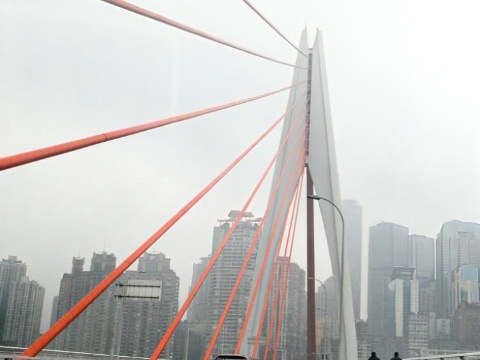}
    \end{subfigure}
    \begin{subfigure}{0.12\hsize}
        \includegraphics[width=\hsize]{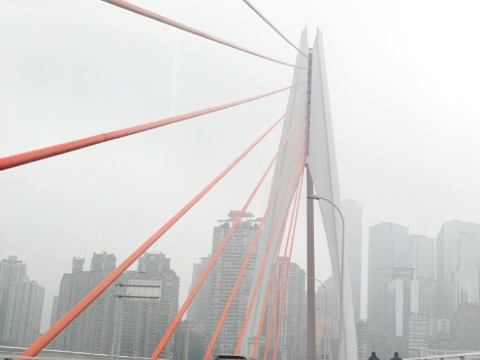}
    \end{subfigure}
    \begin{subfigure}{0.12\hsize}
        \includegraphics[width=\hsize]{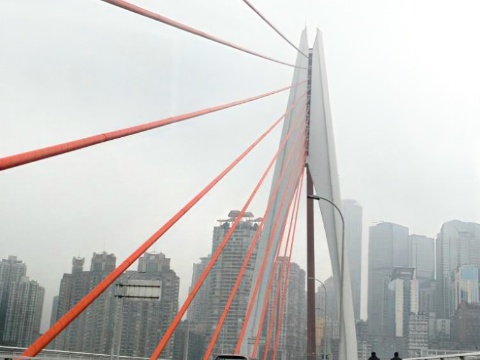}
    \end{subfigure}
    \begin{subfigure}{0.12\hsize}
        \includegraphics[width=\hsize]{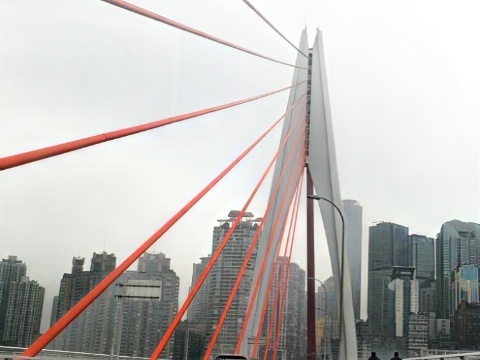}
    \end{subfigure}
    \\
    \begin{subfigure}{0.12\hsize}
        \includegraphics[width=\hsize]{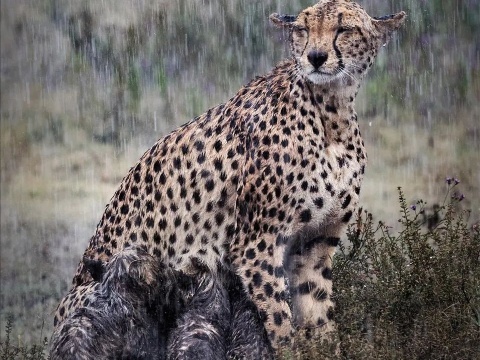}
        \caption{Input}
    \end{subfigure}
    \begin{subfigure}{0.12\hsize}
        \includegraphics[width=\hsize]{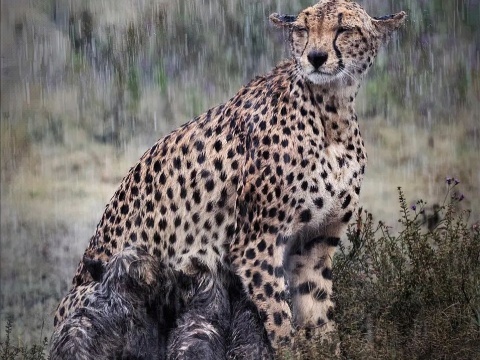}
        \caption{NIMA}
    \end{subfigure}
    \begin{subfigure}{0.12\hsize}
        \includegraphics[width=\hsize]{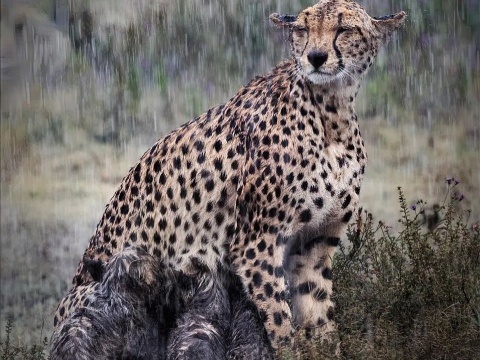}
        \caption{MUSIQ}
    \end{subfigure}
    \begin{subfigure}{0.12\hsize}
        \includegraphics[width=\hsize]{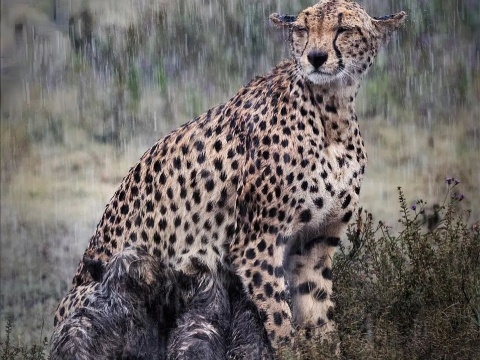}
        \caption{CLIP-IQA}
    \end{subfigure}
    \begin{subfigure}{0.12\hsize}
        \includegraphics[width=\hsize]{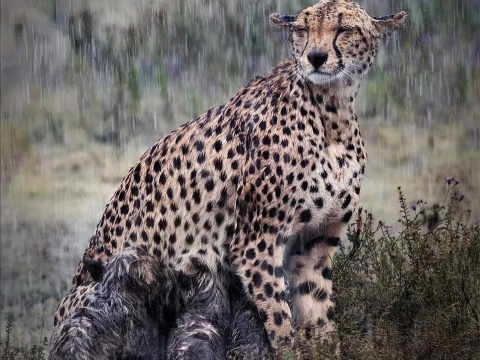}
        \caption{LIQE}
    \end{subfigure}
    \begin{subfigure}{0.12\hsize}
        \includegraphics[width=\hsize]{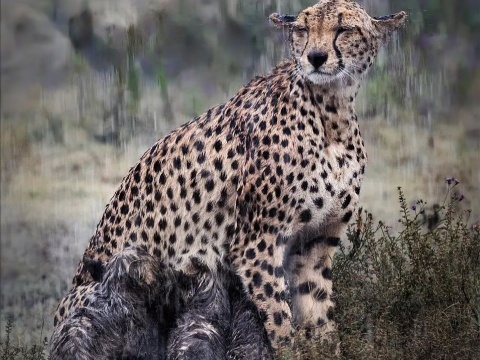}
        \caption{Ours}
    \end{subfigure}
    \\
    \caption{Ablation studies of the proposed image assessment method.}
    \label{fig:vis_ablation-iqa}
\end{figure*}

We conduct experiments to investigate the VLM-based image assessment for selecting pseudo-labels.
We compare our proposed method with existing image quality assessment metrics, including NIMA~\cite{talebi2018nima}, MUSIQ~\cite{ke2021musiq}, CLIP-IQA~\cite{wang2023exploring}, and LIQE~\cite{zhang2023blind}, by replacing the pseudo-label update criteria.
As discussed in \cref{subsubsec:vlmiqa}, our VLM-based rating approach can select pseudo-labels with less weather-related artifacts.
Consequently, the trained models show superior restoration ability, as illustrated in \cref{fig:vis_ablation-iqa}.

\subsubsection{Analysis of semantics regularization.}

\begin{figure}[tp]
    \centering
    \includegraphics[width=1.0\hsize]{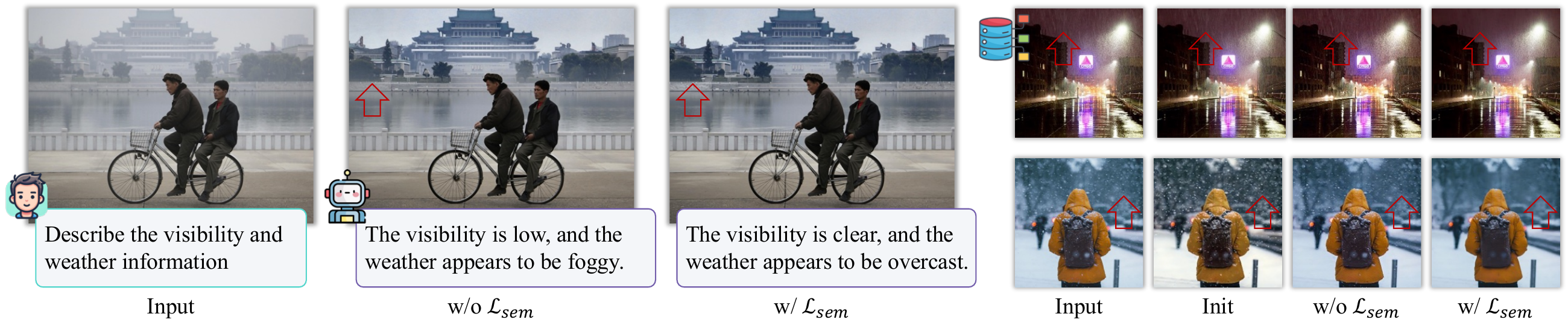}
    \caption{Analysis of the semantics regularization.}
    \label{fig:vis_ablation-sem}
\end{figure}

We study the impact on the semantics regularization based on \cref{fig:vis_ablation-sem}.
The VLM~\cite{chen2024internvl} can detect the nuanced difference of the restored image to be foggy or overcast.
By further monitoring the training process, we observe that the semantics-enhanced approach benefits learning by leading to better-stored pseudo-labels and subsequent training.
Hence, the model trained with the description-assisted semantics regularization $\mathcal{L}_{sem}$ addresses the subtle weather context misalignment, improving the visual quality.

\section{Conclusion}
This paper advances real-world adverse weather image restoration using vision-language models, overcoming the limitations of methods trained on synthetic data.
By evaluating clearness and semantics in natural images, our semi-supervised approach trains models on real, unlabeled images.
Our dual-step strategy, combining image assessment and weather prompt learning, enhances clearness with real data.
Further, semantics enhancement adjusts weather conditions in vision-language model descriptions, addressing context semantics in adverse weather.
Experimental results show that our method outperforms state of the arts. Yet, the computational burden of using large VLMs remains a limitation.


%


\section*{Acknowledgements}
The work was supported by the National Key R\&D Program of China (Grant No. 2022ZD0160100), the Research Grants Council of the Hong Kong Special Administrative Region, China (Grant No. 14201620), and the Hong Kong Innovation and Technology Fund (Grant No. MHP/092/22).

%
%
\bibliographystyle{splncs04}
\bibliography{ref}

\end{document}